%% file: main.tex
\documentclass[journal]{IEEEtranTIE}
\usepackage{amsmath,amssymb,amsfonts}
\usepackage{amsthm,mathrsfs}
\usepackage{algorithmic}
\usepackage[ruled,linesnumbered]{algorithm2e}
\usepackage{graphicx}
\usepackage{textcomp}
\usepackage{xcolor}
\usepackage{multirow}
\usepackage{float}
\usepackage{subfigure}
\usepackage{amssymb}

\usepackage{color}
\usepackage{stfloats}
\usepackage{adjustbox}
\usepackage{subcaption}
\usepackage{multicol}

\newtheorem{theorem}{Theorem}

\usepackage{graphicx}
\usepackage{cite}
\usepackage{picinpar}
\usepackage{amsmath}
\usepackage{url}
\usepackage{flushend}
\usepackage{colortbl}
\usepackage{soul}
\usepackage{multirow}
\usepackage{pifont}
\usepackage{color}
\usepackage{alltt}
\usepackage[hidelinks]{hyperref}
\usepackage{enumerate}
\usepackage{siunitx}
\usepackage{breakurl}
\usepackage{epstopdf}
\usepackage{pbox}

\usepackage{amsmath,amssymb,amsfonts,gensymb,textcomp}
\usepackage{graphicx}
\usepackage[ruled,linesnumbered]{algorithm2e}
\usepackage{xcolor}
\usepackage{multirow}
\usepackage{booktabs}
\usepackage{svg}
\usepackage{romannum}
\usepackage{enumitem}
\usepackage{algorithmic}

\begin{document}

\title{Collaborative State Fusion in Partially Known Multi-agent Environments}
\author{Tianlong Zhou, Jun Shang, Weixiong Rao}

\author{
	\vskip 1em
	
	Tianlong Zhou,	Jun Shang, Weixiong Rao

}

\markboth{Journal of \LaTeX\ Class Files,~Vol.~18, No.~9, September~2020}%
{Collaborative State Fusion in Partially Known Multi-agent Environments}

\maketitle

\begin{abstract}
In this paper, we study the collaborative state fusion problem in a multi-agent environment, where mobile agents collaborate to track movable targets. Due to the limited sensing range and potential errors of on-board sensors, it is necessary to aggregate individual observations to provide target state fusion for better target state estimation. Existing schemes do not perform well due to (1) impractical assumption of the fully known prior target state-space model and (2) observation outliers from individual sensors. To address the issues, we propose a two-stage collaborative fusion framework, namely \underline{L}earnable Weighted R\underline{o}bust \underline{F}usion (\textsf{LoF}). \textsf{LoF} combines a local state estimator (e.g., Kalman Filter) with a learnable weight generator to address the mismatch between the prior state-space model and underlying patterns of moving targets. Moreover, given observation outliers, we develop a time-series soft medoid(TSM) scheme to perform robust fusion. We evaluate \textsf{LoF} in a collaborative detection simulation environment with promising results. In an example setting with 4 agents and 2 targets, \textsf{LoF} leads to a 9.1\% higher fusion gain compared to the state-of-the-art.
\end{abstract}




\input{01-introduction}
\input{02-preliminary}
\input{03-approach}
\input{04-experiments}

\input{05-related-work}
\input{06-conclusion}


\bibliographystyle{IEEEtranTIE}
\bibliography{IEEEabrv, ref}

\begin{IEEEbiography}
[
{
\includegraphics[width=1in,height=1.25in,clip,keepaspectratio]{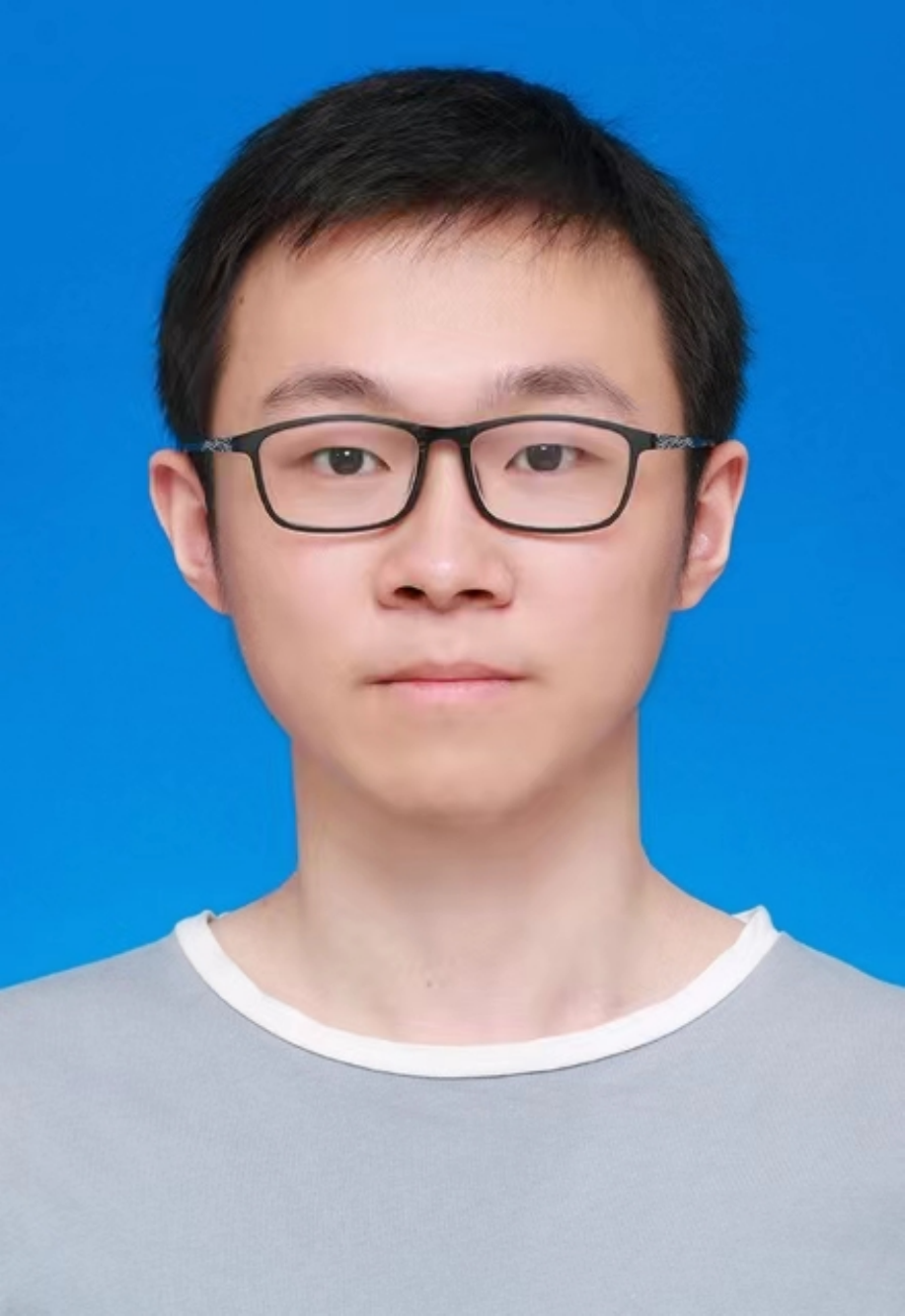}
}]
{Tianlong Zhou}
received the B.E. degree from Tongji University, Shanghai, China, in 2021, where he is currently pursuing the Ph.D. degree with the School of Computer Science and Technology (Software Engineering). His research interests include multi-agent system, reinforcement learning and mobile computing.
\end{IEEEbiography}

\vspace{-4em}

\begin{IEEEbiography}
[
{
\includegraphics[width=1in,height=1.25in,clip,keepaspectratio]{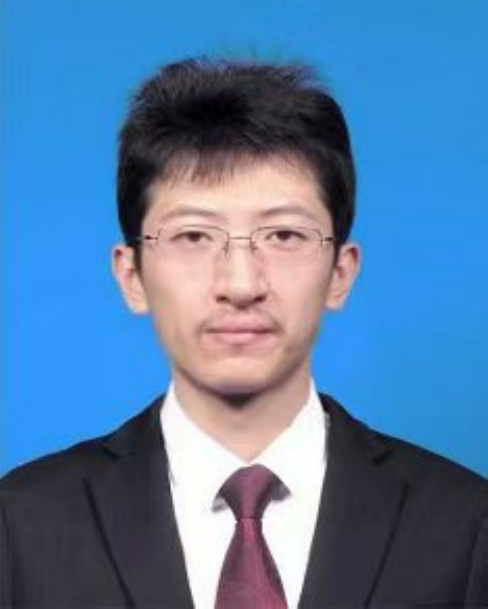}
}]
{Jun Shang} (Senior Member, IEEE) received the B.Eng. degree in control science and engineering from Harbin Institute of Technology, Harbin, China, in 2013, and the Ph.D. degree in control science and engineering from Tsinghua University, Beijing, China, in 2018. 
      
From September 2018 to January 2023, he was a Postdoctoral Fellow with the Department of Electrical and Computer Engineering, University of Alberta, Edmonton, Canada. He is currently a Professor with Tongji University, Shanghai, China.
  
His research interests include fault diagnosis, cyber-physical security, and networked control.
\end{IEEEbiography}

\vspace{-4em}

\begin{IEEEbiography}
[
{
\includegraphics[width=1in,height=1.25in,clip,keepaspectratio]{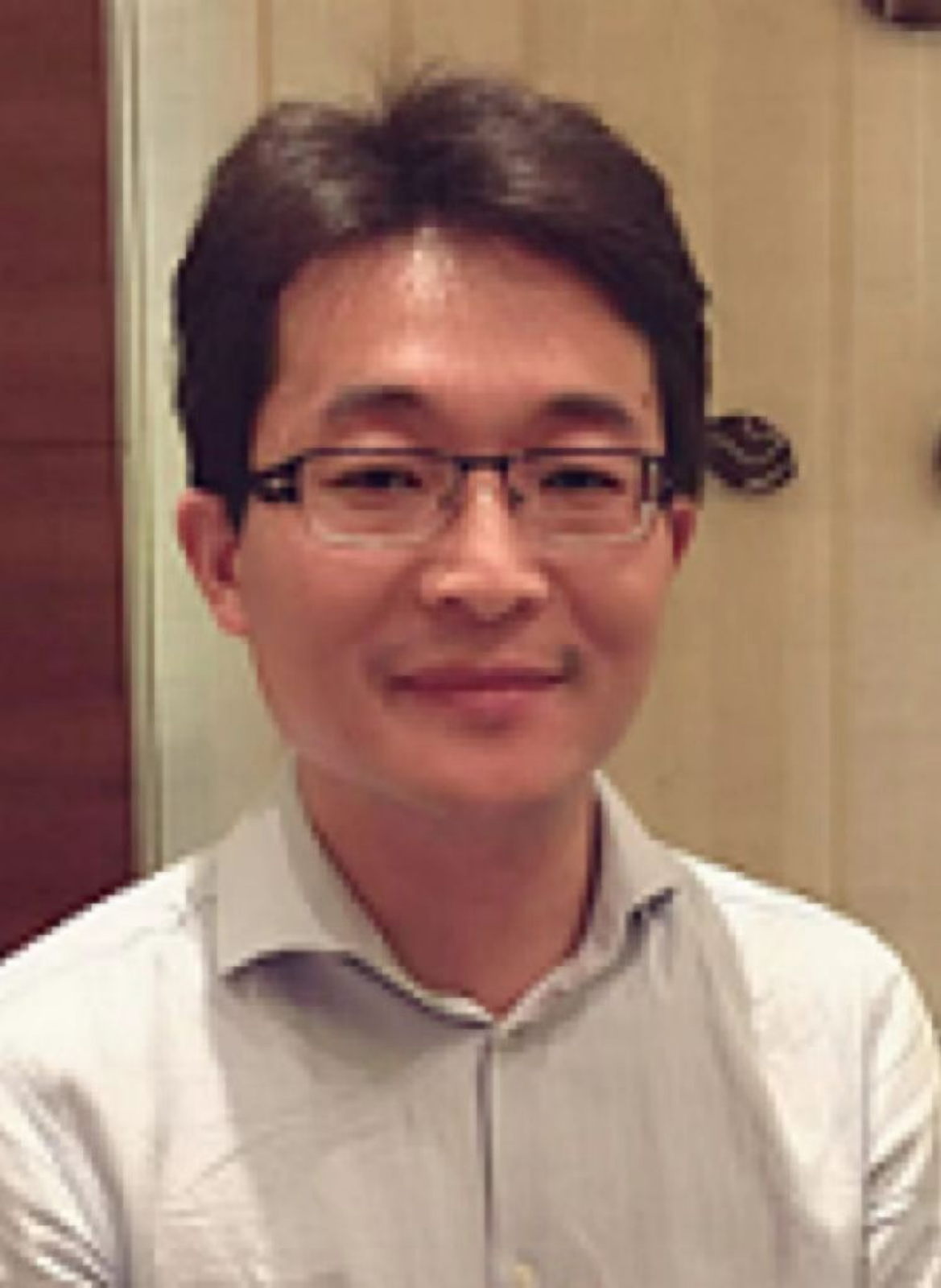}
}
]
{Weixiong Rao}
(Member, IEEE) received the PhD degree from The Chinese University of Hong Kong, Hong Kong, in 2009. After that, he worked for Hong Kong University of Science and Technology(2010), University of Helsinki (2011-2012), and University of Cambridge Computer Laboratory Systems Research Group (2013) as postdoctor researchers. He is currently a full professor with School of Computer Science and Technology (Software Engineering), Tongji University, China (since 2014). His research interests include mobile computing and spatiotemporal data science, and is a member of the CCF and ACM.
\end{IEEEbiography}


\newpage

\input{07-append}

\end{document}

%% file: 01-introduction.tex
\section{Introduction}
In this paper, we study the collaborative fusion problem in a multi-agent environment. 
In Fig. \ref{fig:cd1}, a team of mobile agents collaborates to track movable targets and estimates target states by the observations of onboard range-bearing sensors. 
Due to the limited sensing range and potential errors of on-board sensors, the observations of individual agents are inefficient to provide accurate target state estimates. It is important to aggregate individual observations for target state fusion with better accuracy. This problem has numerous applications, such as search-and-rescue \cite{atif2021uav} and environmental monitoring \cite{yuan2023marine}.

To perform target state fusion, a baseline approach is first to transmit the raw observations from individual agents to a centralized fusion process. The fusion process next aggregates the observations to generate final target state estimates~\cite{zhang2019sequential, kettner2017sequential}. Yet, this centralized baseline approach suffers from high communication cost and computation overhead. Instead, in the distributed fusion scheme~\cite{liu2023bevfusion, tang2023multisensors}, each agent first performs a local light-weighted state estimate and next transmits the local estimate to a centralized fusion process. Since the local estimate is typically with much smaller space size than the raw observation, the distributed fusion leads to trivial communication cost and computation overhead. To perform state fusion, existing approaches frequently assume that the prior target state-space (SS) model is available. When the prior SS model does not match real targets, the distributed approaches, when using \emph{model-based} fusion schemes \cite{sun2016distributed, qu2022probabilistic}, suffer from significant performance degradation. Otherwise, when the target SS model is completely unknown, the distributed approaches, if using the \emph{learning-based} fusion schemes \cite{liu2023bevfusion, chitta2022transfuser} typically with black-box neural networks, cannot provide high interoperability as the model-based schemes do. 

In addition, it is crucial to provide robust state fusion, when either raw observation data or local estimates by individual agents are outliers (e.g., caused by faulty sensors or transmission disturbance). Existing works ~\cite{hein2017formal,wong2018provable} frequently detect disturbance before fusion. {However, these works either require predefined disturbance patterns, or involve non-trivial fine-tuning efforts during the fusion stage, leading to poor generalization capability. A recent work, \emph{Soft Medoid}~\cite{geisler2020reliable}, performs robust fusion to tolerate outlier data, but yet cannot work on the distributed fusion setting where each agent gives a probability distribution of target state estimates.  

\begin{figure}[t]
 \centering
 \includegraphics[width=0.46\columnwidth]{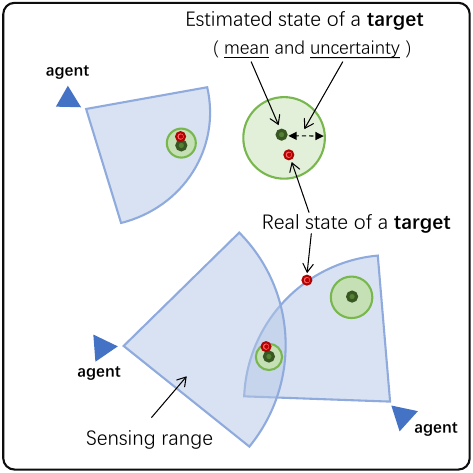}  
 \caption{Collaborative Detection. A team of three agents (blue) exploits range-bearing sensors with limited sensing ranges to track four movable targets (red) by state estimates (green) with the mean and uncertainty.}
 \label{fig:cd1}
 \end{figure}

To address the issues above, we propose a two-stage collaborative fusion framework, namely \underline{L}earnable Weighted R\underline{o}bust \underline{F}usion (\textsf{LoF}). Specifically, unlike either the fully known or completely unknown SS model, \textsf{LoF} performs in the more general setting with \emph{partially} known SS model, where the prior SS model only roughly matches the real target patterns. Given the partially known setting, \textsf{LoF} leverages the local estimates (typically with a probability distribution of state estimates) by a state estimator, e.g., either the classic Kalman Filter or the recent work WalNut \cite{zhou2024learning}, to learn the weight of such a local estimate. That is, differing from the previous works on state estimation, \textsf{LoF} offers the learned weight to identify the importance of the local estimate. After receiving the local estimates and weights from individual agents, the centralized process performs robust fusion via the developed \underline{T}ime series \underline{S}oft \underline{M}edoid (TSM). It performs robust fusion on top of target state probability distribution. Moreover, TSM is adaptive to the change of target state estimates over time. This enables robust fusion of the target state estimates from  collaborative multi-agents with partially known SS models. We make the following contribution in this paper.
\begin{itemize}
    \item We propose a two-stage collaborative state fusion framework \textsf{LoF} on partially known state-space models to provide high interpretability and robustness. 
    \item We develop a time-series soft medoid scheme (TSM) to perform robust fusion on the state probability distribution.
    \item We evaluate \textsf{LoF} in a collaborative detection simulation environment with promising results. In an example setting with 4 agents and 2 targets, \textsf{LoF} leads to a 9.1\% higher fusion gain compared to the state-of-the-art BCI~\cite{sun2016distributed}.
\end{itemize}
The rest of this paper is organized as follows. Section \ref{sec2} first introduces related works and 
preliminaries and Section \ref{sec3} then gives problem definition. Next, Section \ref{sec4} presents solution details. After that, Section \ref{sec5} evaluates our work.
Finally, Section \ref{sec7} concludes the paper. Table \ref{tab:notation} lists the mainly used symbols and meanings in this paper.

\begin{table}[htbp]
\begin{center}\footnotesize
\caption{Mainly used symbols and meaning\\ (obs. observation; cov. covariance. )}\label{tab:notation}\vspace{-1ex}
\begin{tabular}{clcl}
\toprule
\footnotesize Symbol & \footnotesize Meaning & \footnotesize Symbol & \footnotesize Meaning \\
 \midrule
 $I, J$ & \footnotesize  Num. of agents/targets & $\Delta \mathbf{y}_{t}^{i}$ & \footnotesize Innovation \\
  $i, j$ & \footnotesize Agent and target ID & $\mathbf{S}_{t|t-1}^{i}$ & \footnotesize Innovation cov.  \\
 $\hat{x}_{t}^{i}$/$\mathbf{x}_{t}$ & \footnotesize Local/Fused state mean & $\mathbf{H}$  & \footnotesize Obs. matrix \\ $\hat{\mathit{\Sigma}}_{t}^{i}$/$\mathbf{\Sigma}_{t}$ & \footnotesize Local/Fused state cov. & $\mathbf{R}$ & \footnotesize Obs. noise cov. \\
${N}_{t}^{i}$/$\mathcal{N}_{t}$ & \footnotesize Local/Fused Gaussian dist.  & $J_{t}(\cdot)$ & \footnotesize JS divergence \\
$w_{t}^{i}$/$\mathbf{w}_{t}^{i}$  & \footnotesize Local/Fused weight & $\mathbf{r}_{t}^{i}$ & \footnotesize Balancing coefficient \\
\bottomrule
\end{tabular}
\end{center}
\end{table}
\normalsize
\vspace{-2em}


%% file: 02-preliminary.tex
\section{Related Work and Preliminaries}\label{sec2}

\subsection{Related Work}
\textbf{Sensor Fusion} aims to provide state estimates with higher accuracy than those obtained from individual sensors. In general, depending upon whether or not raw measurement data are directly used for fusion, literature works are classified into two categories including \textit{centralized fusion} and \textit{distributed fusion}. By effectively leveraging all available raw measurements, {centralized fusion} typically leads to excellent estimation in certain scenarios~\cite{zhang2019sequential, ullah2017hierarchical}. In contrast, {distributed fusion} employs multiple nodes to process raw measurements, thereby significantly reducing computational load and communication costs while also safeguarding individual privacy~\cite{fan2022distributed, qu2022probabilistic}. Moreover, according to whether or not the state space model is available as prior knowledge, the {distributed fusion} can be further divided into three following manners.

\emph{(1) Model-based approaches} fuse observations frequently under the Bayesian framework requiring the availability of prior SS models. For example, the work~\cite{sun2016distributed} exploits the well-known Batch Covariance Intersection scheme to enable the distributed fusion. Instead,~\cite{qu2022probabilistic} investigates the fusion problem with probabilistic constraints in the time-varying system. Though with high Interpretability, these approaches perform heavily depending upon the completely accurate prior state-space (SS) model, and do not well when the true target SS model mismatches the prior one. 

\emph{(2) Learning-based approaches} develop end-to-end neural network (NN)-based fusion. For example, the work~\cite{liu2023bevfusion} proposes the Encoder-Decoder based NN by concatenating multimodal observation encodings for fusion, and~\cite{chitta2022transfuser} proposes a transformer NN to fuse the encoded multimodal observations. Though without the prior state-space model, these works require significant training data and typically perform not well with few label data. 

\emph{(3) Hybrid approaches} aim to provide 
better performance and interoperability than the approaches above. VINFNet~\cite{tang2023multisensors} performs state estimation with a learnable parameterized state-space model and fusion scheme. Unlike our work, yet without state estimation uncertainties. The work \cite{gao2021multi} learns the state and uncertainty of individual sensors, and next performs the fusion over the learned uncertainties. Yet, this previous work is motivated by the correlation between targets, and the state estimates of individual sensors are completely independent of the state-space model, resulting in low accuracy. Instead, we focus on 
robust fusion in a partially known state-space model with the focus on an interpretable fusion framework with the mean and uncertainty of state estimates.

\textbf{Robust Fusion} involves two following lines of literature works. Firstly, the works \cite{hein2017formal, entezari2020all} perform explicit disturbance detection before robust fusion. For instance, \cite{entezari2020all} models the relationship of data points as a graph and next leverages the adjacency matrix of the graph to detect perturbations. However, these works typically perform well on predefined perturbation patterns with low generalization to alternative patterns. To overcome these issues, some emerging works focus on self-adjusting the fusion process~\cite{zhu2019robust, geisler2020reliable}. For example, \cite{zhu2019robust} learns data uncertainty in a data-driven manner and performs robust fusion based on the learned uncertainty. Yet, this work is restricted to non-temporal data with no interpretability. For better interpretability, \cite{wu2020graph} introduces an information-theoretic loss function to mitigate perturbations but with no uncertainty fusion. \emph{Soft Medoid}~\cite{geisler2020reliable} introduces a differentiable aggregation scheme for deep learning, significantly improving fusion robustness. Nevertheless, Soft Medoid \emph{(i)} works only for point data, instead of time series data, and \emph{(ii)} fails to address uncertainty-based fusion, which is vital in collaborative detection. To tackle these challenges, we propose a variant of Soft Medoid, namely TSM, for robust fusion on uncertain time series data.

\textbf{Collaborative Detection} has many real-world applications. For example, in UAV-based parcel delivery~\cite{kuo2024uav}, targets are usually static, with the primarily research focus on scalability~\cite{blais2024scalable}, robustness~\cite{li2020robust}, and communication challenges~\cite{blumenkamp2021emergence}. In contrast, search and rescue~\cite{wang2023cooperative} involves dynamic targets, where limited sensing capabilities make it rather hard to obtain precise state information of the targets, leading to uncertainties in the target's state. To address the challenges, some collaborative detection systems study robust decision-making algorithm with help of target state estimate~\cite{hsu2021scalable, zhou2024learning}. The works~\cite{hsu2021scalable} achieve promising results by assuming complete known SS models. Instead, the work~\cite{zhou2024learning} introduces a more practical scenario where the SS model is only partially known. Yet, this work does not provide robust fusion, which is yet our focus of this paper.

\subsection{Preliminaries}

\subsubsection{State Space Model}
A typical \emph{State Space} (SS) model of a dynamic system consists of the state-evolution and observation models.

\begin{equation*}\footnotesize
\begin{array}{lllllll}
    \mathbf{x}_{t}&=&f( \mathbf{x}_{t-1} )+\mathbf{u}_{t},\quad & \mathbf{u}_{t} &\sim& {N}(\mathbf{0}, \mathbf{Q}) & \mathbf{x}_{t} \in \mathbb{R}^{m} \\    \mathbf{y}_{t}&=&h(\mathbf{x}_{t})+\mathbf{v}_{t},\quad & \mathbf{v}_{t} &\sim& {N}(\mathbf{0}, \mathbf{R}) & \mathbf{y}_{t} \in \mathbb{R}^{n}
\end{array}
\end{equation*}

In the equation above, $\mathbf{x_t}$ is the latent state vector of the system at time step $t$. It evolves from the previous state $\mathbf{x}_{t-1}$ by a state-evolution function $f(\cdot)$ and an additive white Gaussian noise variable $\mathbf{u}_{t}$ with a covariance matrix $\mathbf{Q}$. Next, $\mathbf{y}_{t}$ is the observation over the latent state $\mathbf{x_t}$ by an observation function $h(\cdot)$ and an additive white Gaussian noise $\mathbf{v}_{t}$ with a covariance matrix $\mathbf{R}$.

\begin{equation*}\small
\begin{array}{lll}
    \mathbf{x}_{t}&=&\mathbf{F} \mathbf{x}_{t-1} +\mathbf{u}_{t} \\
    \mathbf{y}_{t}&=&\mathbf{H} \mathbf{x}_{t}+\mathbf{v}_{t}
\end{array}
\label{eq:linear-ss}
\end{equation*}

For a linear SS model, $\mathbf{F}$ and $\mathbf{H}$ indicate the state evolution and observation matrix, respectively. For a partially known SS setting, we assume that the elements in $\mathbf{F}$, $\mathbf{H}$, $\mathbf{Q}$ and $\mathbf{R}$ might be incorrect. For instance, the work~\cite{revach2022kalmannet} employ a rotation matrix to make the rotated elements deviate from correct ones.

{For a collaborative detection application, e.g., on a 2D map, studied in this paper, $\mathbf{x}_{t}$ indicates the (real) location and velocity of targets, and $\mathbf{y}_{t}$ is a pair consisting of Euclidean distance and bearing (azimuth), observed by range-bearing sensors equipped on mobile agents.} Due to the limited sensing range of these sensors, we have observation samples when the target is within the range of the sensors, and otherwise empty samples. When the sensors suffer from faults, such samples may contain outliers.

\textit{Kalman Filter} (KF). For a linear Gaussian SS model, KF recursively estimates the system state distribution $p(\mathbf{x}_{t}|\mathbf{y}_{1:t})$ using the previous estimation $p(\mathbf{x}_{t-1}|\mathbf{y}_{1:t-1})$ and the new observation $\mathbf{y}_{t}$ by the following prediction and update steps.

\begin{align*}\footnotesize
p(\mathbf{x}_{t}|\mathbf{y}_{1:t-1}) &=\int p(\mathbf{x}_{t}|\mathbf{x}_{t-1})p(\mathbf{x}_{t-1}|\mathbf{y}_{1:t-1}) \mathrm{d}\mathbf{x}_{t-1} \\
p(\mathbf{x}_{t}|\mathbf{y}_{1:t}) &= \frac{p(\mathbf{y}_{t}|\mathbf{x}_{t})p(\mathbf{x}_{t}|\mathbf{y}_{1:t-1})}{\int p(\mathbf{y}_{t}|\mathbf{x}_{t})p(\mathbf{x}_{t}|\mathbf{y}_{1:t-1})\mathrm{d}\mathbf{x}_{t}} 
\end{align*}

where the 1st equation above is the prediction step of KF, and the 2nd one is the update step. Here, $p(\mathbf{x}_{t}|\mathbf{x}_{t-1})$ is the state-evolution function, and $p(\mathbf{y}_{t}|\mathbf{x}_{t})$ is the likelihood of $\mathbf{y}_{t}$.

\textit{Model-based State Estimation}. With the help of the state-space model, Bayesian Filters (such as KF) can recursively estimate the system state $\mathbf{x}_{t}$ by utilizing the effective part of observation $\mathbf{y}_{t}$. Specifically, KF will calculate the innovation $\Delta \mathbf{y}_{t}=\mathbf{y}_{t}-\hat{\mathbf{y}}_{t}$ and associate covariance matrix (i.e. uncertainty) $\mathbf{S}_{t|t-1}$. Intuitively, the innovation represents the information from observations that can be used to estimate the target. The covariance $\mathbf{S}_{t|t-1}$ represents the degree of confidence in that information. The KF uses $\mathbf{S}_{t|t-1}$ as a weight to select a portion of the information in the innovation for estimating the target state.

\subsubsection{Soft Medoid}\label{sec2.2}
To understand the developed TSM scheme, we give
an introduction of \textit{Soft Medoid} ~\cite{geisler2020reliable}. With the purpose to address the challenge that weighted averages can be distorted arbitrarily by outliers, {Soft Medoid} mitigates the impact of outliers while ensuring robust fusion performance. Specifically, it interpolates between the \emph{weighted medoid} and the \emph{weighted average}. The weighted medoid, a multivariate generalization of the weighted median, effectively reduces the impact of perturbations but at the cost of fusion performance. In contrast, the weighted average guarantees fusion performance but is more susceptible to perturbations. When given $I$ data points $x_{i}$ and weights $\mathbf{w}_{i}$, \textit{Soft Medoid} offers a balanced compromise between the two approaches to have fusion result $\Bar{\mathbf{x}}$.

\vspace{-1ex}
\footnotesize
\begin{equation*}
\Bar{\mathbf{x}} = \frac{\sum_{i=1}^{I}{(r_{i}\mathbf{w}_{i}x_{i})}}{\sum_{i=1}^{I} r_{i}\mathbf{w}_{i}}, \quad
 r_{i}=\frac{\mathsf{exp}(-\frac{1}{\mathsf{T}}\sum_{k}^{I}\mathbf{w}_{k}||x_{i}, x_{k}||)}{\sum_{k=1}^{I}\mathsf{exp}(-\frac{1}{\mathsf{T}}\sum_{j}^{I}\mathbf{w}_{j}||x_{j}, x_{k}||)} 
\end{equation*}\vspace{-2ex}
\normalsize

In the equations above, the denominator $\sum_{i=1}^{I} r_{i}\mathbf{w}_{i}$ within $\Bar{\mathbf{x}}$ is to normalize the numerator $\sum_{i=1}^{I}{(r_{i}\mathbf{w}_{i}x_{i})}$, and {the balancing coefficient $r_{i}$} involves the trade-off between the weighted medoid and the weighted average, by a temperature $\mathsf{T}$. That is, for $\mathsf{T}\to \infty$, all data points  converge to the same values of $r_{i}$ and $\Bar{\mathbf{x}}$ then becomes a {weighted average} with $\Bar{\mathbf{x}} = \frac{\sum_{i=1}^{I} \mathbf{w}_i x_i}{\sum_{i=1}^{I}\mathbf{w}_i}$. Otherwise, for $\mathsf{T}\to 0$, {we have one data point with $r_{i}\to 1$ and all others with $r_i\to 0$. The soft medoid $\Bar{\mathbf{x}}$ then becomes a weighted medoid $\Bar{\mathbf{x}} = arg \min_{x_i} \sum_{k=1}^{I} \mathbf{w}_{k} ||x_i, x_k||$, where $||\cdot||$ indicates Euclidean distance. The data point with $r_{i}\to 1$ is also known as the weighted medoid.} 

{Thanks to the fully differentiable property of Soft Medoid ~\cite{geisler2020reliable}, it is rather comfortable to incorporate Soft Medoid into a neural network and next build a learnable model.}

\section{Problem Formulation}\label{sec3}
Given $I$ mobile agents (sensors) and $J$ movable targets, we denote the observation sequence with the horizon length $H$ 
by $\mathcal{Y} = \{ \mathcal{Y}^{j}_{t} \}|_{1\leq t\leq H,1\leq j\leq J}$. Here, $\mathcal{Y}_{t}^{j} = \{ \mathbf{y}_{t}^{i,j}\}|_{1\leq i\leq I}$ indicates $I$ measurements on target $j$ at time step $t$ by the agents $i=1...I$. When target $j$ is out of the sensing range of agent $i$ or the measurement sample is lost, the sample $\mathbf{y}_{t}^{i,j}$ is empty. We thus have the observation $\mathbf{y}_{t} \in \mathbb{R}^{n} \cup\left\{*\right\}$, where ``*" means an empty sample. Next, we denote the ground truth target states at time step $t$ by $\mathcal{X}_t = \{ \mathbf{x}^{j}_{t}\}|_{j = 1...J}$, where $\mathbf{x}^{j}_{t}$ is the real state of target $j$ at time step $t$.

In this paper, we study the collaborative state fusion problem to estimate the probability $ p( \mathcal{X}_{t}| \mathcal{Y}_{1:t} )$. {Given the measurement samples $\mathcal{Y}_{1:t}$ observed by $I$ agents, we aggregate such samples to estimate the target state probability distribution $ p( \mathcal{X}_{t}| \mathcal{Y}_{1:t} )$. By the classic work~\cite{deng2012sequential}, we assume that each target state of $j$ is independent and follows a multivariate Gaussian distribution. Thus, for a certain target $j$, we are interested in the probability} $ p( \mathbf{x}_{t}^{j}| \mathcal{Y}_{1:t}^{j} )$ with the Gaussian distribution $\mathcal{N}(\tilde{\mathbf{x}}_{t}^{j}, \tilde{\mathbf{\Sigma}}_{t}^{j})$, where $\tilde{\mathbf{x}}_{t}^{j}$ and $\tilde{\mathbf{\Sigma}}_{t}^{j}$ are the mean and covariance matrix, respectively. For example, $\tilde{\mathbf{x}}_{t}^{j}$ indicates the point estimation of target state (e.g., location or speed), and $\tilde{\mathbf{\Sigma}}_{t}^{j}$ can be treated as the uncertainty of state estimation. With this uncertainty-based state estimation, it is helpful for the downstream application, such as collaborative detection, to
make robust decision, e.g., the movement (directions or speeds) of mobile agents, by a multi-agent reinforcement learning algorithm \cite{zhou2024learning}.

For simplicity, we remove the index $j$ from the variables $\mathbf{x}_{t}^{j}$ and $\mathcal{Y}_{1:t}^{j}$ (due to the independence of target states) and have the simple notation $p( \mathbf{x}_{t}| \mathcal{Y}_{1:t} )$ for state estimation.


%% file: 03-approach.tex
\section{Solution Detail}\label{sec4}

\subsection{Overall Framework}\label{sec:3A}
In this section, we first give overall framework of \textsf{LoF}, and next present the detail of two key components. Given the problem definition above, a centralized baseline solution may send all raw observation samples from $I$ agents to a centralized process and suffer from high communication and computation overhead. To overcome this issue, we develop a two-stage fusion framework. Each agent $i$ with $1\leq i\leq I$ first performs \emph{local estimation}, then sends local estimates to a centralized process, which next conducts \emph{robust fusion}. After that, the fusion result is available for each agent for better local estimation. Moreover, in the collaborative detection scenario, each agent might take the fusion state (with the state estimate mean and uncertainty) as the input to its decision maker for the control of moving directions and speeds.

{We show that our two-stage solution can work correctly, mainly due to Theorem 1 (the proof refers to the appendix). 

\begin{theorem}\label{them1}
We solve the collaborative state estimation by

\begin{subequations}\footnotesize
    \begin{align}    \underbrace{p(\mathbf{x}_{t}|\mathcal{Y}_{1:t})}_{ \mathcal{N}(       \hat{\mathbf{x}}_{t}, \hat{\mathbf{\Sigma}}_{t})        }
         &= \sum_{i=1}^{I}(
        {\underbrace{        p(\mathbf{x}_{t}|\mathcal{Y}_{1:t-1}, \mathbf{y}_{t}^{i})} 
        _{N(\hat{x}_{t}^{i}, \hat{\mathit{\Sigma}}_{t}^{i}) }
        }
        \cdot \underbrace{p(i|\mathcal{Y}_{1:t})}_{\mathbf{w}_{t}^{i}}) \label{eq:center-fuse} \\
        \underbrace{p(\mathbf{x}_{t}|\mathcal{Y}_{1:t-1},  \mathbf{y}_{t}^{i})}_{N(\hat{x}_{t}^{i}, \hat{\mathit{\Sigma}}_{t}^{i}) }
 & = \psi (
        \underbrace{ p(\mathbf{x}_{t-1}|\mathcal{Y}_{1:t-1})}_{\mathcal{N}(
        \hat{\mathbf{x}}_{t-1}, \hat{\mathbf{\Sigma}}_{t-1} )       
        }, \enspace \mathbf{y}_{t}^{i}
        ) \label{eq:local-estimate} \\
\underbrace{p(i|\mathcal{Y}_{1:t})}_{\mathbf{w}_{t}^i}       
         & = \phi (\underbrace{p(\mathbf{y}_{t}^{i}|\mathcal{Y}_{1:t-1})\cdot p(i|\mathcal{Y}_{1:t-1})}_{w_{t}^{i}}) \label{eq:lweight}
     \end{align}
\end{subequations}

\label{th:fusion-problem}
\end{theorem}
\vspace{-1em}

In Eq. (\ref{eq:center-fuse}}), the fusion state $p(\mathbf{x}_{t}|\mathcal{Y}_{1:t})$ is the weighted sum of \emph{local state estimates} $p(\mathbf{x}_{t}|\mathcal{Y}_{1:t-1}, \mathbf{y}_{t}^{i})$ by the \emph{fusion weights} $\mathbf{w}^i_t=p(i|\mathcal{Y}_{1:t})$. 
To generate the {fusion weights}, we require the local weights $w_{t}^{i}$ from $I$ agents and then develop a local state estimator $\psi(\cdot )$. Next in Eq. (\ref{eq:local-estimate}), the estimator $\psi(\cdot )$ takes as input the observation $\mathbf{y}_{t}^{i}$ provided by agent $i$ and the previous fusion state $p(\mathbf{x}_{t-1}|\mathcal{Y}_{1:t-1})$. 
Given the Gaussian distribution of state estimates, we denote the fusion state (resp. local state) by $\mathcal{N}(\hat{\mathbf{x}}_{t}, \hat{\mathbf{\Sigma}}_{t})$ with the mean $\hat{\mathbf{x}}_{t}$ and covariance $\hat{\mathbf{\Sigma}}_{t}$ (resp. 
$N(\hat{x}_{t}^{i}, \hat{\mathit{\Sigma}}_{t}^{i})$ with the 
 mean $\hat{x}_{t}^{i}$ and covariance $\hat{\mathit{\Sigma}}_{t}^{i}$). 

After the local weights and local estimates are transmitted from mobile agents, the centralized fusion process generates the {fusion weight} $\mathbf{w}^i_t$ in Eq. (\ref{eq:lweight}) via the normalization function $\phi(\cdot)$ over these \emph{local weights} $w_t^i=p(\mathbf{y}_{t}^{i}|\mathcal{Y}_{1:t-1})\cdot p(i|\mathcal{Y}_{1:t-1})$. From this equation, we find that the fusion weight $\mathbf{w}_{t}^{i}$ at time step $t$ actually performs the recursive normalization over the likelihood of historical observation, i.e., $p(\mathbf{y}_{t}^{i}|\mathcal{Y}_{1:t-1})$, and 
the previous fusion weight $\mathbf{w}_{t-1}^{i}=p(i|\mathcal{Y}_{1:t-1})$, at time step $t-1$. Here, this likelihood indicates how the local observation  $\mathbf{y}_{t}^{i}$ provided by agent $i$ is likely to the history one $\mathcal{Y}_{1:t-1}$. A smaller value of the likelihood means a large discrepancy between the current observation and history one. The local weight $w_{t}^{i}$ then becomes smaller, and vice versa.

\begin{figure}[h]
\centering
\includegraphics[width=.92
\linewidth]{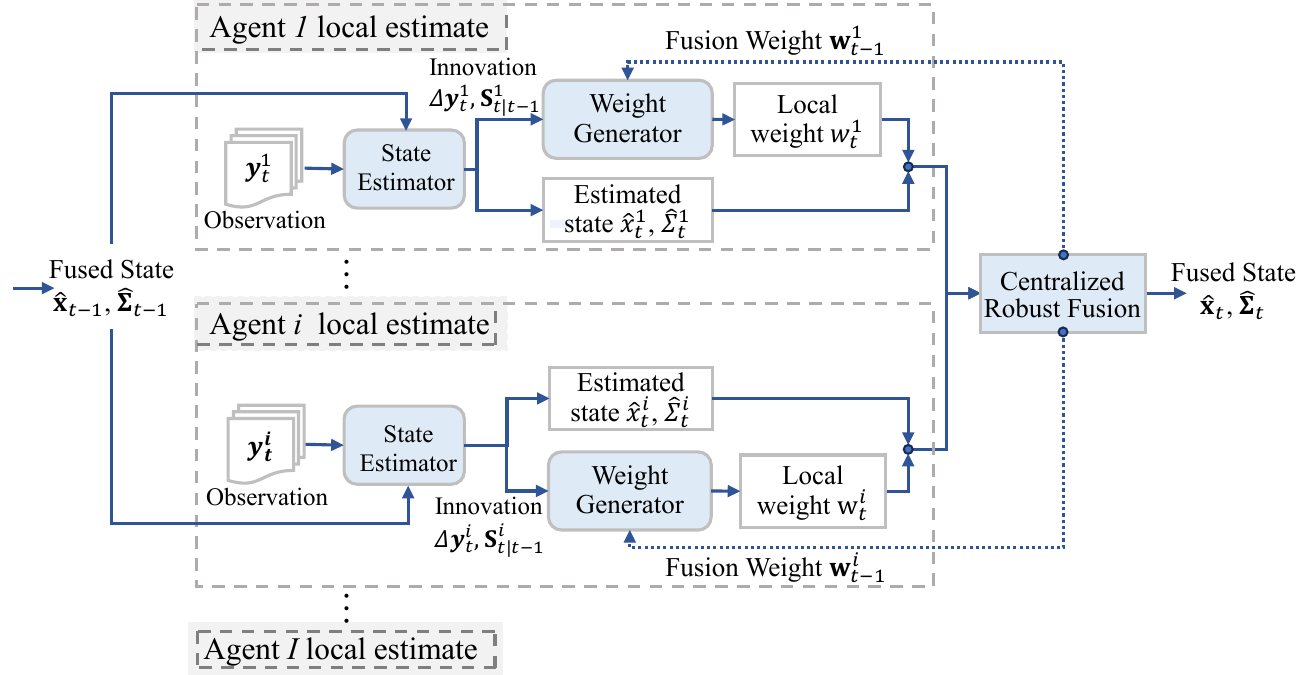} 
\caption{Overall framework}\label{fig:framework}
\vspace{-1ex}
\end{figure}

Following Theorem \ref{them1}, Fig.~\ref{fig:framework} gives the collaborative fusion framework with two stages of \emph{local estimation} and \emph{robust fusion}. Each agent $i$ takes its current observation $\mathbf{y}_{t}^{i}$ together with the previous fusion state $\mathbf{\hat{x}}_{t-1}, \mathbf{\hat{\Sigma}}_{t-1}$ and the fusion weight $\mathbf{w}^i_{t-1}$ (both can be queried from the centralized fusion process) as input, and generates a local state estimate $\psi(\cdot)$ and local weight $w^i_t$. When receiving the local state estimates and weights from all agents, the centralized process first generates the fusion weight $\mathbf{w}_t^i$ via the normalization function $\phi(\cdot)$, and performs robust fusion to generate the fusion state $p(\mathbf{x}_{t}|\mathcal{Y}_{1:t})$ by the current mean and covariance $\mathbf{\hat{x}}_{t}, \mathbf{\hat{\Sigma}}_{t}$. 

When the SS model of targets is fully known, we may explicitly obtain the local estimation $\psi(\cdot)$ and local weight $w^i_t$ ~\cite{masnadi2019step}. Otherwise, given a partially known SS model, the accurate estimation of $\psi(\cdot)$ and $w^i_t$ is hard. To this end, on imperfect local estimates, our fusion solution can still provide accurate fusion state via a data-driven local weight generator that will be given soon.

\vspace{-1em}

\subsection{Local Estimation}

\subsubsection{Local State Estimator}
\label{sec:state-estimator}
Each agent $i$ exploits the classic Kalman Filter (KF) or the recent work WalNut \cite{zhou2024learning} as the local estimator $\psi(\cdot )$ to generate the local state estimate of $p(\mathbf{x}_{t}|\mathcal{Y}_{1:t-1},   \mathbf{y}_{t}^{i})$. 
The input data of the estimator $\psi(\cdot )$ include the previous time-step fusion state $p(\mathbf{x}_{t-1}|\mathcal{Y}_{1:t-1})$ and current observation $\mathbf{y}_{t}^{i}$.
The output of $\psi(\cdot )$ includes the local mean $\hat{x}_{t}^{i}$ and covariance $\hat{\mathit{\Sigma}}_{t}^{i}$ of the Gaussian distribution $N(\cdot)$ as the local state estimate. Moreover, the KF-like estimator $\psi(\cdot )$ gives the innovation $\Delta\mathbf{y}_{t}^{i} = \mathbf{y}_{t}^{i}-\hat{\mathbf{y}}_{t}^{i}$ and uncertainty $\mathbf{S}_{t|t-1}^i$ (that are used to generate the local weight $w_{t}^{i}$).

\begin{equation}\small
    N(\hat{x}_{t}^{i}, \hat{\mathit{\Sigma}}_{t}^{i}), \enspace\Delta\mathbf{y}_{t}^{i}, \enspace  \mathbf{S}_{t|t-1}^{i} = \mathsf{KF}(\hat{\mathbf{x}}_{t-1}^{i}, \hat{\mathbf{\Sigma}}_{t-1}^{i}, \mathbf{y}_{t}^{i})
\end{equation}

\subsubsection{Local Weight Generator}
\label{sec:clg}
To generate the local weight $w_t^i=p(\mathbf{y}_{t}^{i}|\mathcal{Y}_{1:t-1})\cdot p(i|\mathcal{Y}_{1:t-1})$ in Eq. (\ref{eq:lweight}), 
we first give the general idea. Here, the 2nd subitem 
$p(i|\mathcal{Y}_{1:t-1})$ is just the fusion weight 
$\mathbf{w}_{t-1}^i$ of the previous time step $t-1$. Since the fusion weight can be queried from the centralized fusion process, we focus on the 1st subitem $p(\mathbf{y}_{t}^{i}|\mathcal{Y}_{1:t-1})$, i.e., the likelihood of the current local observation $\mathbf{y}_{t}^{i}$ over the history observations $\mathcal{Y}_{1:t-1}$.

\begin{theorem}\label{th:o-weight}
The observation likelihood is computed by
\begin{equation}\label{eq:o-weight-cal}
    p(\mathbf{y}_{t}^{i}|\mathcal{Y}_{1:t-1})  = \mathsf{PDF}(\Delta \mathbf{y}_{t}^{i}|\langle \mathbf{0},\mathbf{S}_{t|t-1}^{i}\rangle)
\end{equation}
\end{theorem}
The right-hand item of Eq. (\ref{eq:o-weight-cal}) indicates that the probability of generating the innovation $\Delta \mathbf{y}_{t}^{i} = \mathbf{y}_{t}^{i}-\hat{\mathbf{y}}_{t}^{i}$ by the probability density function (PDF) of the Gaussian distribution with the mean $\mathbf{0}$ and covariance $\mathbf{S}_{t|t-1}^{i}$, under the KF framework. Intuitively, a higher value of $p(\mathbf{y}_{t}^{i}|\mathcal{Y}_{1:t-1})$ means that the observation $\mathbf{y}_{t}^{i}$ by agent $i$ is more likely to be consistent with the history observations $\mathcal{Y}_{1:t-1}$.

\begin{figure}[t]
\centering
\includegraphics[width=0.8\linewidth]{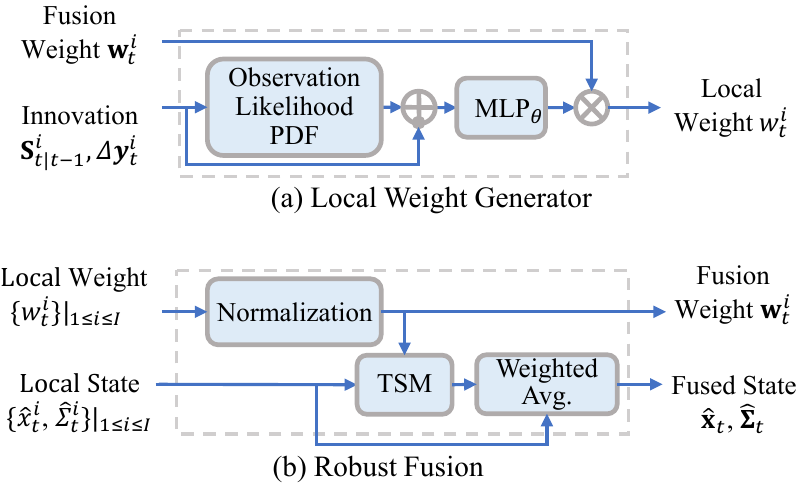} 
\caption{(a) Local Weight Generator, (b) Robust Fusion}
\vspace{-1ex}
\label{fig:confgen}
\end{figure}

Due to the partially known SS model, 
the PDF function in Theorem \ref{th:o-weight} may not work well to accurately estimate the likelihood $p(\mathbf{y}_{t}^{i}|\mathcal{Y}_{1:t-1})$. To this end, in Fig. \ref{fig:confgen}(a), we develop a Multilayer Perceptron (MLP) to learn the likelihood in a data-driven fashion. The MLP involves two input items: the probability generated by the $\textsf{PDF}(\cdot)$ in Eq. (\ref{eq:o-weight-cal}), and the innovation $\Delta \mathbf{y}_{t}^{i}$. Using the light weighted MLP is mainly due to the online weight generation requiring trivial overhead. 
Next, we follow Eq. (\ref{eq:lweight}) to generates the local weight $w_{t}^{i}$. 

\begin{equation}\small
w_{t}^{i} = \underbrace{\mathtt{MLP}_{\theta}(\textsf{PDF}(\Delta \mathbf{y}_{t}^{i}|\langle \mathbf{0}, \mathbf{S}_{t|t-1}^{i}\rangle),\Delta \mathbf{y}_{t}^{i})}_{p(\mathbf{y}_{t}^{i}|\mathcal{Y}_{1:t-1})} \cdot \mathbf{w}_{t-1}^{i}
\label{eq:learned-likelihood}
\end{equation}

In Eq. (\ref{eq:learned-likelihood}), $\mathtt{MLP}_{\theta}$ requires the innovation $\Delta\mathbf{y}_{t}^{i} = \mathbf{y}_{t}^{i}-\hat{\mathbf{y}}_{t}^{i}$ as input. Here,  when observation samples are lost during transmission, $\mathbf{y}_{t}^{i}$, i.e., the local observation by agent $i$, may be missing and the innovation $\Delta\mathbf{y}_{t}^{i}$ is then unavailable. Thus, we are inspired by~\cite{kordestani2020new}: the likelihood of missing observations can be successfully approximated by a large value of $\Delta \mathbf{y}_{t}^{i}$. Thus, since the observation likelihood follows a Gaussian distribution of the mean $\mathbf{0}$ and covariance matrix $\mathbf{S}_{t|t-1}^{i}$, we choose a great value of $\Delta \mathbf{y}_{t}^{i}$ that is outside the $2\sigma$ range of this Gaussian distribution \cite{yan2012state}. Note that the $2\sigma$ range works on a univariate Gaussian distribution. In our multivariate setting, we choose the Mahalanobis distance $D_{M}$=$[(\Delta \mathbf{y}_{t}^{i})^{T}(\mathbf{S}_{t|t-1}^{i})^{-1}(\Delta \mathbf{y}_{t}^{i})]^{\frac{1}{2}}$ $\ge 2$ to set the value of $\Delta \mathbf{y}_{t}^{i}$~\cite{yan2012state}.

\vspace{-1em}
\subsection{Robust Fusion}\label{sec:rf}
In this section, we first give a baseline fusion approach and next propose a robust fusion solution.

\subsubsection{Baseline Approach}
As shown in Fig.~\ref{fig:confgen}(b), the centralized fusion involves two steps: generating fusion weights and performing robust state estimation. Firstly, given the local weights $w_t^i$ from $I$ agents, the centralized fusion generates 
the fusion weights $\mathbf{w}_{t}^i$ via the normalization function $\phi (\cdot)$.
\vspace{-1ex}
\begin{equation}\small
    \mathbf{w}_{t}^{i} = \frac{w_{t}^{i}}{\sum_{k=1}^{I}w_{t}^{k}}\label{eq13}
\end{equation}\vspace{-1ex}

Secondly, following Eq. (\ref{eq:center-fuse}), we estimate the fusion state as the weighted sum of local states. Due to the Gaussian distribution of fusion and local states, we comfortably estimate the fusion state $\mathcal{N}_t(\hat{\mathbf{x}}_{t}, \hat{\mathbf{\Sigma}}_{t})$ over the local ones $N_t^{i}(\hat{x}_{t}^{i}, \hat{\mathit{\Sigma}}_{t}^{i})$ by ${\mathcal{N}}_{t} = \sum_{i=1}^{I}{(\mathbf{w}_{t}^{i} \cdot {N}_{t}^{i}})$. Since a linear combination of Gaussian distributions is still a Gaussian distribution, we can compute the mean and covariance of $\hat{\mathcal{N}}_{t}$.

\begin{equation}\footnotesize
\hat{\mathbf{x}}_{t} = \sum_{i=1}^{I}{\left(\hat{x}_{t}^{i}\cdot \mathbf{w}_{t}^{i}\right)}, \enspace
\hat{\mathbf{\Sigma}}_{t} = \sum_{i=1}^{I}{\left[ \hat{\mathit{\Sigma}}_{t}^{i}+(\hat{x}_{t}^{i}-\hat{\mathbf{x}}_{t})(\hat{x}_{t}^{i}-\hat{\mathbf{x}}_{t})^{\mathsf{T}}\right]\cdot \mathbf{w}_{t}^{i}}
\label{eq14}
\end{equation}

\subsubsection{Robust Fusion Solution}
When the local estimates involve outliers, e.g., caused by either faulty agent sensors or transmission errors, the baseline approach above may not work well: the fusion weights or states above could be distorted arbitrarily by outliers.

To overcome the issue above, we propose a robust fusion scheme. Recall that the {Soft Medoid}~\cite{geisler2020reliable} is originally developed to compromise between weighted medoid and weighted average on a set of $I$ data points $x_i$. Yet, in our setting, beyond the \textit{point estimates} at the current time step $t$, the fusion process may involve \emph{time series estimates} of multiple time steps $1,...,t$. Thus, we develop the variant, \textit{Time Series Soft Medoid} (TSM), by the adaptation to time series data by exploiting a distance $D_t(\cdot)$ to replace the original Euclidean one $||x_i, x_k||$ in Section \ref{sec2.2}. The distance $D_t(\cdot)$ introduces the Jensen-Shannon divergence  $J_t(i,k)$ at time $t$ between the state estimate distributions of agents $i$ and $k$.


\begin{equation}\footnotesize
D_{t}(k,i) =  D_{t-1}(i,k) + \tau_{t}^{i,k} \cdot \left(J_{t}(i,k) -D_{t-1}(i,k)\right)
\label{eq:tsm-dist}
\end{equation}

The distance $D_{t}(\cdot)$ above involves the trade-off between the current JS divergence $J_{t}(\cdot)$ and previous distance $D_{t-1}(\cdot)$ by an adaptive decay factor {$\tau_{t}^{i,k}\in [0.0, 1.0]$}. {Unlike the Euclidean distance in the original Soft Medoid, the developed distance $D_{t}(\cdot)$ exploits the temporal and spatial patterns of the state estimate distributions of agents $i$ and $k$. That is, we first use the Jensen-Shannon divergence to identify the spatial distance of the distributions, and next incorporate the effect of temporal pattern by a decay factor $\tau_{t}^{i,k}$.

\textit{Adaptive Decay Factor}: Following the work~\cite{noda2010adaptation}, we employ the 2nd order derivative of $D_t(\cdot)$ to tune the decay factor $\tau_{t}^{i,k}$, such that the distance $D_t(\cdot)$ is adaptive to the aforementioned trade-off. {Yet, simply using the 2nd order derivative may lead to the decay factor $\tau_{t}^{i,k}$ out of the range $[0.0, 1.0]$. Moreover, as shown in the work \cite{noda2010adaptation}, the change trend of $\tau_{t}^{i,k}$ is very sharp near 0.5 and yet rather smooth near 0.0 or 1.0. Thus, we apply the Sigmoid function to match such a trend and  meanwhile ensure $\tau_{t}^{i,k}\in [0.0, 1.0]$.}

\begin{subequations}\footnotesize
\begin{align}
\tau_{t}^{k,i} & = \textsf{sigmoid}(\hat{\tau}_{t}^{k,i}) \\
\hat{\tau}_{t}^{k,i} &= \hat{\tau}_{t-1}^{k,i} + \gamma \cdot\frac{\mathrm{d}^{2}}{\mathrm{d} t^{2}}  D_{t}(k,i)\\
  \frac{\mathrm{d}^{2} }{\mathrm{d} t^{2}} D_{t}(k,i)&= \frac{\mathrm{d}}{\mathrm{d} t} D_{t}(k,i)- \frac{\mathrm{d} }{\mathrm{d} t}J_{t-1}(k,i)
\end{align}\label{eq8}
\end{subequations}
We follow the work~\cite{bui2024newton} to set the learning rate $\gamma=10^{-3}$ by default. Here we give the benefit of the \textsf{sigmoid} function over $\hat{\tau}_{t}^{k,i}$ as follows. If the growth trend of the 2nd order derivative of $D_t(\cdot)$ in Eq. (\ref{eq8}b) is consistent with the underlying dynamic patterns of target states, the \textsf{sigmoid} function works correctly to match such patterns. Instead, the 2nd order derivative of $D_t(\cdot)$ is rather abnormal (caused by the faulty sensors) even with a very large outlier of $D_t(\cdot)$, the \textsf{sigmoid} function has chance to clip this outlier and makes robust fusion. In this way, we have chance to overcome the issue that the 2nd order derivative of $D_t(\cdot)$ is a great value even out of the range $[0.0, 1.0]$ caused by the outliers of the distance $D_t(\cdot)$. 

\subsection{Training} \label{sec34}
To ensure that our two-stage fusion framework \textsf{LoF} works, we need to learn the network parameters $\theta$ of $\mathtt{MLP}_{\theta}$ that is used to predict the observation likelihood and next generate local weights $w_t^i$ in Eq. (\ref{eq:learned-likelihood}). With help of downstream collaborative detection applications, we may collect the dataset $\mathcal{D}$ with $|\mathcal{D}|$ trajectories $\tau = (\mathcal{Y}_{1...H}, \mathcal{X}_{1...H})$ with the horizon length $H$. Here, the ground-truth target state $\mathcal{X}$ can be as the labels for the observation $\mathcal{Y}$. Yet, this dataset does not directly provide the labels of the observation likelihood and thus can not explicitly learn the parameters ${\theta}$ of $\mathtt{MLP}_{\theta}$. 

Fortunately, we note that the fusion state $\mathcal{N}_t(\hat{\mathbf{x}}_{t}, \hat{\mathbf{\Sigma}}_{t})$ depends on the fusion weight $\mathbf{w}_{t}$, which depends on the local weight $w_t$ that further depends on the observation likelihood $\mathcal{N}_t(\hat{\mathbf{x}}_{t}, \hat{\mathbf{\Sigma}}_{t}) \rightarrow \mathbf{w}_{t} \rightarrow  w_t \rightarrow \mathtt{MLP}_{\theta} \rightarrow  \mathbf{y}_{t}$. By the \emph{chain rule}, we can comfortably learn the parameters $\theta$ of $\mathtt{MLP}_{\theta}$ via the gradient back-propagation algorithm on the dataset $\mathcal{D}$ above. Thus, we develop the \textit{n}egative \textit{l}og \textit{l}ikelihood (NLL) loss~\cite{qian2022uncertainty}, mainly because the fusion state involves a Gaussian distribution $\mathcal{N}_t(\hat{\mathbf{x}}_{t}, \hat{\mathbf{\Sigma}}_{t})$, rather than the mean-based point value fusion. Due to the Gaussian distribution, the NLL loss involves the mean and covariance. We then have the following NLL loss, where $\det\left(\cdot\right)$ indicates the matrix determinant.

\begin{equation}\footnotesize
\begin{split} 
\label{eq:loss-LoF-total}  
\mathcal{L}\left(\theta \right) &= \frac{1}{H} \cdot\sum_{t=1}^{H} {\left(\log \det( \hat{\mathbf{\Sigma}}_{t} )+
{\left (\mathbf{x}_{t}- \hat{\mathbf{x}}_{t}\right )}^{\mathsf{T}} {\hat{\mathbf{\Sigma}}_{t}^{-1}}
{\left (\mathbf{x}_{t}- \hat{\mathbf{x}}_{t}\right ) }\right)}
\end{split}
\end{equation}

Learning the parameters $\theta$ requires a differentiable function. Note that the original Soft Medoid is differentiable with respect to the fusion weight $\mathbf{w}_{i}$. The proposed variant TSM only modifies the distance metric, but with no change of its differentiability. Thus, the chain rule $\mathcal{N}_t(\hat{\mathbf{x}}_{t}, \hat{\mathbf{\Sigma}}_{t}) \rightarrow \mathbf{w}_{t}$  in TSM is still valid to learn the parameter $\theta$.

\begin{algorithm}\footnotesize
    \textbf{Input:} Dataset $\mathcal{D}$, parameter $\theta$, batch size $B$, horizon length $H$\;
    \For{each iteration}{
        Sample $D$ trajectories from $\mathcal{D}$\;
        \For{each trajectory $\tau$}{
            \For{each time step $t$}{
                \For{each agent $i$}{
                   Est. local state $N(\hat{x}_{t}^{i}, \hat{\mathit{\Sigma}}_{t}^{i},\Delta\mathbf{y}_{t}^{i}, \mathbf{S}_{t|t-1}^{i}$ (\S \ref{sec:state-estimator})\;
                    Est. local weight $w_{t}^{i}$  (\S \ref{sec:clg})\;
                }
                Estimate $\mathcal{N}(       \mathbf{x}_{t}, \mathbf{\Sigma}_{t})$ by robust fusion (\S~\ref{sec:rf})\;
            }
            Compute NLL loss $loss_{\tau} =\mathcal{L}\left(\theta \right)$  (\S \ref{sec34}).
        }
        Update $\theta$ with $\frac{1}{D}\sum_{\tau=1}^{D}loss_{\tau}$ by Adam Optimizer.
    }
    \caption{Train \textsf{LoF} Model}
\label{alg:WALNUT}
\end{algorithm}

%% file: 04-experiments.tex
\section{Experiments}\label{sec5}
\subsection{Experimental Setup}\label{exp:setting}
We perform the experiments for the following objective.
\begin{itemize}
    \item How does \textsf{LoF} perform in partially known SS settings?
    \item How does \textsf{LoF} work well to tolerate perturbations?
    \item What benefit can \textsf{LoF} offer for downstream application?
\end{itemize}
To this end, we develop a multi-agent reinforcement learning (MARL) collaborative detection environment on top TTenv~\cite{ttenv} to evaluate our work \textsf{LoF} with the focus on the partially known SS model.


\subsubsection{Target State-Evolution Model} The ground truth of target states, denoted by $\mathbf{x}_{t}={[ \mathbf{x}_{t,1}, \mathbf{x}_{t,2}, \dot{\mathbf{x}}_{t,1}, \dot{\mathbf{x}}_{t,2} ]}^T$, involves the coordinates ($\mathbf{x}_{t,1}, \mathbf{x}_{t,2}$) and velocity $(\dot{\mathbf{x}}_{t,1}, \dot{\mathbf{x}}_{t,2})$ within a 2D map. Targets move by  a linear state evolution model.

\begin{equation*}\footnotesize
\begin{array}{clcl}
    \mathbf{x}_{t+1} & =  \mathbf{A}\mathbf{x}_{t} + u_{t}, \enspace
    u_{t}  \sim  \mathcal{N}(0, \mathbf{Q}); 
    &\mathbf{A} =&\left[\begin{array}{ll}
    \mathbf{I}_{2} & \tau_{t} \mathbf{R}_{\alpha} \mathbf{I}_{2} \\
    0 & \mathbf{I}_{2}    \end{array}\right]\\
    \mathbf{R}_{\alpha} & = \left[\begin{array}{ll}
        \cos \alpha & -\sin \alpha \\
        \sin \alpha & \cos \alpha
        \end{array}\right]; 
    &\mathbf{Q} =&\left[\begin{array}{ll}
        \tau_{t}^{3} / 3 \mathbf{I}_{2} & \tau_{t}^{2} / 2 \mathbf{I}_{2} \\
        \tau_{t}^{2} / 2 \mathbf{I}_{2} & \tau_{t} \mathbf{I}_{2}
        \end{array}\right]  
\end{array}
\end{equation*}

In the above equations, $\mathbf{A}$ is the target state-evolution matrix, $\tau_{t}$ is the time interval between two neighboring time steps, $\mathbf{R}_{\alpha}$ is a rotation matrix, $\mathbf{Q}$ is the covariance matrix of the additional noise, $u_{t}$ is the noise variable, and $\mathbf{I}_{2}$ is a $2 \times 2$ identity matrix. Without loss of generality, we define the \emph{partially known} model by setting the parameter $\alpha \in [0\degree, 360\degree]$ where $\alpha=0\degree$ indicates the fully known state evolution model and otherwise a partially known model \cite{revach2022kalmannet}. 

\subsubsection{Agent observation model} We define the agent observation by $\mathbf{a}_{t} = [ \mathbf{a}_{t,1}, \mathbf{a}_{t,2}, \mathbf{a}_{t,\theta} ]$, where the first two items indicate the 2D coordinates and the third one the azimuth. By a range-bearing sensor, an agent identifies targets within the range by the following observation model~\cite{ttenv}:

\footnotesize
\begin{equation*}\vspace{-2ex}
    \mathbf{y}_{t} = h(\mathbf{a}_{t},\mathbf{x}_{t})+\mathbf{v}_{t},\quad \mathbf{v}_{t} \sim {N}(\mathbf{0}, \mathbf{R}) 
\end{equation*}\normalsize

where

\scriptsize 
\begin{equation*}
h(\mathbf{a}, \mathbf{x})=
\left[\begin{array}{l} r_{\mathbf{a}, \mathbf{x}}:=\sqrt{\left(\mathbf{x}_1-\mathbf{a}_1\right)^2+\left(\mathbf{x}_2-\mathbf{a}_2\right)^2} \\
\phi_{\mathbf{a}, \mathbf{x}}:=\tan ^{-1}((\mathbf{x}_2-\mathbf{a}_2)(\mathbf{x}_1-\mathbf{a}_1))-\mathbf{a}_\theta + \beta
\end{array}\right], 
\mathbf{R}=\rho \left[\begin{array}{ll}
        \sigma_{r}^{2} & 0 \\
        0 & \sigma_{b}^{2}
        \end{array}\right]
\end{equation*}\normalsize

Here, the range-bearing sensor within an agent measures the Euclidean distance $r_{\mathbf{a}, \mathbf{x}}$ and bearing (azimuth) $\phi_{\mathbf{a}, \mathbf{x}}$ between targets and agents. Moreover, the measurement noise involves the standard deviations of the distance and bearing, i.e., $\sigma_{r}$ and $\sigma_{b}$, respectively. 
Following the work~\cite{revach2022kalmannet}, we define the \emph{partially known observation} model by the rotation angle $\beta$, where $\beta=0^\circ$ indicates the fully known observation model and otherwise a partial one. 
{The parameter $\rho$ ($\ge 0$) controls the strength of measurement noise $\mathbf{v}_t$. A larger $\rho$ corresponds to stronger noise, while $\rho = 0.0$ indicates no noise. By default, we set $\rho = 1.0$.}

\subsubsection{Perturbation Setting}\label{sec5.13}  Following the work~\cite{meng2022resilient}, we inject the bias $[1.0\ m, 0.1\ rad]$ to the raw observation $\mathbf{y}_{t}$. That is, we first randomly choose one from all $I$ agents and perturb the raw observation of the chosen agent at a predefined time step $t=20$. It simulates the scenario where the sensor of this chosen agent suffers from a permanent failure and cannot be recovered afterwards.

\subsubsection{Counterpart Competitors}
\label{exp:counter-com}
\begin{itemize}
    \item Batch Covariance Intersection (BCI)~\cite{sun2016distributed} is a model-based fusion method. Both our work \textsf{LoF} and BCI first perform local estimation via Kalman filtering and next send the estimates for centralized fusion. Yet, BCI simply uses the covariance matrix as the fusion weight, requiring intensive matrix computation cost. The comparison of BCI and \textsf{LoF} evaluates which one is more effective in a partially known SS setting.
    \item Sequential Kalman Filter~\cite{kettner2017sequential} (SKF) is a classic model-based centralized fusion method. That is, all observation samples are transmitted to a centralized fusion process, which exploits SKF to fuse all transmitted samples, leading to high communication and computation cost. By comparing SKF and \textsf{LoF}, we study the performance between the centralized and collaborative fusion schemes.
    \item The recent work \cite{gao2021multi} proposes a hybrid approach with the model-based and learning-based fusion (in short Hyb). Unlike \textsf{LoF} that tunes fusion weights by neural networks (NNs), Hyb exploits NNs as state estimators and then performs fusion estimates by BCI. When the state-space model is fully unknown, this Hyb approach has been shown to be effective. In this paper with the partially known SS model, we compare which performs better between Hyb and \textsf{LoF}.
\end{itemize}

\subsubsection{Metrics}\label{sec:metric}
We develop the evaluation metrics for state estimation (including estimation errors and uncertainty quantification) and collaborative detection, respectively.

\begin{itemize}
    \item \textit{Mean Squared Error (MSE)} in decibels, namely, \textit{MSE}[dB] = 10 $\times\log_{10}$ \textit{MSE}. Thus, a smaller dB value of \textit{MSE}[dB] indicates better state estimation. 
    \item \textit{Fusion Gain (FG)} defines the improvement ratio between the state estimation with and without fusion. Denote $e_{i}$ to be the estimate MSE of each agent $i$, and $e$ the fusion MSE. We  compute  \emph{FG} = $\frac{ \sum_{i=1}^{M}e_{i} - e\cdot M }{\sum_{i=1}^{M}e_{i}}  \times 100\%$. A greater \emph{FG} indicates better performance.
    
        \item \textit{Mean Negative Log-Likelihood} (\textit{MNLL}): the uncertainty by the likelihood $-\frac{1}{T}\sum_{t=1}^{T}{\log \mathcal{N}( \mathbf{x}_{t};\mathbf{\hat{x}}_{t},\mathbf{\hat{\Sigma}}_{t})}$~\cite{qian2022uncertainty}. A smaller \textit{MNLL} means less uncertainty with stable result.
    \item {\textit{Detection Ratio} represents the proportion of targets that are accurately detected by mobile agents. Let $b_{t}^{i}$ be a binary indicator where $b_{t}^{i}=1$ if the condition $| \mathbf{x}{t}^{i} - \hat{\mathbf{x}}{t}^{i} |\le th$ is met, and 0 otherwise. With a default threshold of $th=0.646$, the detection ratio is computed as $R=\frac{1}{T \times M}\sum_{t=1}^{T} \sum_{i=1}^{M}b_{t}^{i}$~\cite{zhou2024learning}. A higher $R$ means better detection accuracy.}
    
\end{itemize}

We conduct experiments on a server with an Intel Xeon Platinum 8255C processor running at 2.50GHz, 40GB of RAM, and the Ubuntu 18.04 operating system. Table \ref{tab:key-para} lists key  (hyper-)  parameters in the evaluation.

\begin{table}[H]
\footnotesize
\begin{center}
\caption{Key (Hyper-) Parameters}\vspace{-1ex}
\label{tab:key-para}
\begin{tabular}{lll}
\toprule
\textbf{Type}                                     & \textbf{Parameter}                         & \textbf{Default Val. and Range}    \\
\midrule
\multirow{3}{*}{Environment}             & Horizon Length $H$                      &40, $-$                \\
                                         & Time Interval                &0.5s, $-$          \\
                                         & Map Size                & $30.0\times30.0 [m^{2}]$, $-$                    \\
                                         \midrule
\multirow{2}{*}{Target}                  & Max Speed               & $2 m/s$, $-$                     \\
                                         & Num. of Targets $J$                 &2, $[0,100]$               \\ \midrule
\multirow{3}{*}{Agent}                   & Max Speed                  & 2 $m/s$, $-$                  \\
                                         & Num. of Agents $I$                &4, $[0,100]$                  \\
                                         & Field of View            & $100 {\degree}$, $[0{\degree}, 360 {\degree}]$              \\ \midrule
\multirow{4}{*}{SS model}                        & $\alpha$ of rotation matrix &$20{\degree}$, $[0{\degree}, 360 {\degree})$                 \\ 
            & $\beta$ of obs. matrix &$10{\degree}$, $[0{\degree}, 360 {\degree})$                 \\ 
            & $\rho$ of obs. noise &$1.0$, $-$                   \\ 
            & $\sigma_{range}$ of obs. noise &$0.2 [m]$ $-$                  \\ 
            & $\sigma_{bearing}$ of obs. noise &$0.01 [rad]$, $-$                  \\ 
\midrule
Decay factor & Learning rate $\gamma$ &$0.001$, $-$                  \\ 
\midrule
\multirow{4}{*}{Training} & Dataset Size  $|\mathcal{D}|$                 &$1300$, $-$                \\
                                         & Iterations                            &$500$, $-$                 \\
                                         & Learning rate                    &$0.003$, $-$                  \\
                                         & Batch size                        & 16, $-$                \\
                                         \bottomrule
\end{tabular}
\end{center}
\end{table}

\subsection{Evaluation Result}
\subsubsection{Baseline Study} Table~\ref{exp:performance-CD} compares the performance of four counterparts with various numbers of targets and agents. Here, `4a2t' indicates 4 agents and 2 targets. From this table, we have the following findings. (\emph{i}) Our work \textsf{LoF} outperforms all four methods in all three settings. For instance,  \textsf{LoF} achieves the smallest MSE[dB] -11.40, highest fusion gain (FG) 87.6\% and the least MNLL 1.26 in the `4a2t' setting. (\emph{ii}) Hyb performs worst among the four approaches. It is mainly because the learning-based estimator does not work well in the partially known SS setting, consistent with the work~\cite{revach2022kalmannet}. Meanwhile, the model-based fusion BCI performs better than SKF in the partially known environment. (\emph{iii}) The MSE values of all approaches in the `4a2t' setting is better than those in the `2a4t' setting, meanwhile with the greater FG values. It makes sense because  more agents lead to more local estimates to fusion and thus have chance to result in better fusion results. In the rest of this section, since we focus on the fusion problem, we use the `4a2t' setting by default to evaluate the fusion performance.  

\begin{table}[htbp]
\begin{center}\footnotesize
\caption{Baseline study.}\label{exp:performance-CD}\vspace{-1ex}
\begin{tabular}{llccc}
\toprule
& & \multicolumn{2}{c}{State} & \multicolumn{1}{c}{Uncertainty}  \\ \cmidrule(lr){3-4}\cmidrule(lr){5-5}
& & MSE[dB]$\downarrow$ & FG(\%) $\uparrow$& MNLL$\downarrow$ \\ \hline
\multirow{4}{*}{4a2t} & BCI \cite{sun2016distributed} & -9.01 ± 0.18          & 78.5 ± 0.6          & 307.90 ± 1.24        \\
&SKF \cite{kettner2017sequential}      &     -5.03 ± 0.51       &  46.2 ± 2.4         &   256.40 ± 2.56   \\
&Hyb \cite{gao2021multi}           & -4.14 ± 0.51          & 34.1 ± 3.0          & 32.88 ± 2.33    \\
&LoF          &\textbf{-11.40 ± 0.06} & \textbf{87.6 ± 0.7} & \textbf{1.26 ± 0.11} \\
\midrule
\multirow{4}{*}{4a4t} & BCI \cite{sun2016distributed} & -6.36 ± 0.09          & 74.8 ± 0.1         & 262.01 ± 0.13     \\
&SKF \cite{kettner2017sequential}      &  -3.38 ± 0.16   &  50.0 ± 0.8         &   233.28 ± 0.01     \\
&Hyb  \cite{gao2021multi}         & -0.90 ± 0.25          & 11.3 ± 3.3         & 55.13 ± 1.59    \\
&LoF          &\textbf{-6.95 ± 0.13} & \textbf{78.0 ± 0.2} & \textbf{1.41 ± 0.07} \\
\midrule
\multirow{4}{*}{2a4t} & BCI \cite{sun2016distributed} & -2.73 ± 0.08          & 57.8 ± 0.1        & 187.69 ± 0.68    \\
&SKF \cite{kettner2017sequential}       &  -1.36 ± 0.23  &  42.0 ± 1.6     &   179.48 ± 1.19     \\
&Hyb \cite{gao2021multi}            & -0.62 ± 0.19          & 31.4 ± 1.4          & 25.64 ± 1.18    \\
&LoF          &\textbf{-3.23 ± 0.11} & \textbf{62.3 ± 1.9} & \textbf{6.44 ± 0.32} \\
\bottomrule
\end{tabular}
\end{center}
\vspace{-1.5em}
\end{table}
\normalsize

\subsubsection{Effect of Partially Known State Space Setting}
We first vary the rotation angle $\beta$ to study the effect of a \textit{partially known observation model}. In Fig.~\ref{fig:exp-pkom}, when $\beta$ becomes greater until $10\degree$ indicating the more inaccurate observation model, \textsf{LoF} outperforms three competitors with the least MSE error and smallest MNLL (uncertainty). Yet, those model-based and hybrid approaches BCI and Hyb do not work well in this partially known observation model.

Next, Fig.~\ref{fig:exp-pkem} evaluates the performance in a \textit{partially known state-evolution model} by setting the parameter $\alpha={0^\circ, 10^\circ, 20^\circ}$. Here, to mitigate the impact of the partially known observation model, we 
particularly set $\beta=0^\circ$. As shown in this figure, when the value of $\alpha$ grows, \textsf{LoF} exhibits the comparable MSE as BCI and SKF. Nevertheless, \textsf{LoF} still performs best with the least MNLL to provide a strong capability of uncertainty representation.

{In Fig. \ref{fig:exp-mn}, We also examine the effect of \emph{measurement noise} by setting the parameter $\rho $ equal to 0.5, 1.0 and 2.0. Here, a higher value of $\rho$ indicates stronger noise and vice versa. In this figure, the MSE errors of \textsf{LoF} on the three values of $\rho$ remain much more stable than three alternatives, and performs the best when $\rho=2.0$, demonstrating that \textsf{LoF} is robust towards the noise.

As a summary, \textsf{LoF} performs rather well in the partially known SS environment. The adopted negative log-likelihood loss function allows \textsf{LoF} to achieve higher-quality uncertainty estimates, while the local weights based on the innovation can effectively improve the overall fusion performance particularly in partially known observation models.

\begin{figure}
  \centering
  \begin{minipage}{0.48\columnwidth}
    \centering
    \includegraphics[width=.95\linewidth]{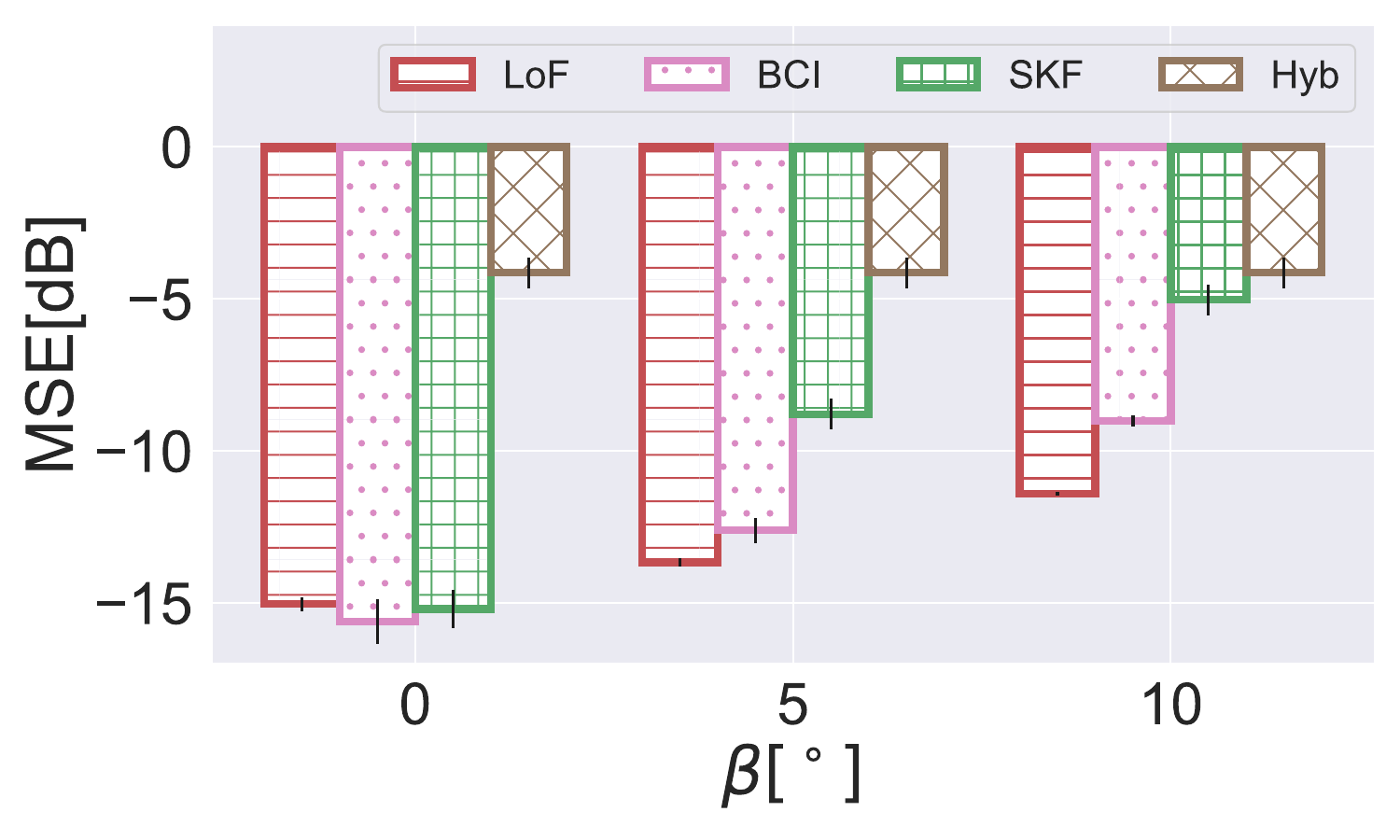}
  \end{minipage}
  \begin{minipage}{0.48\columnwidth}
    \centering
    \includegraphics[width=.95\linewidth]{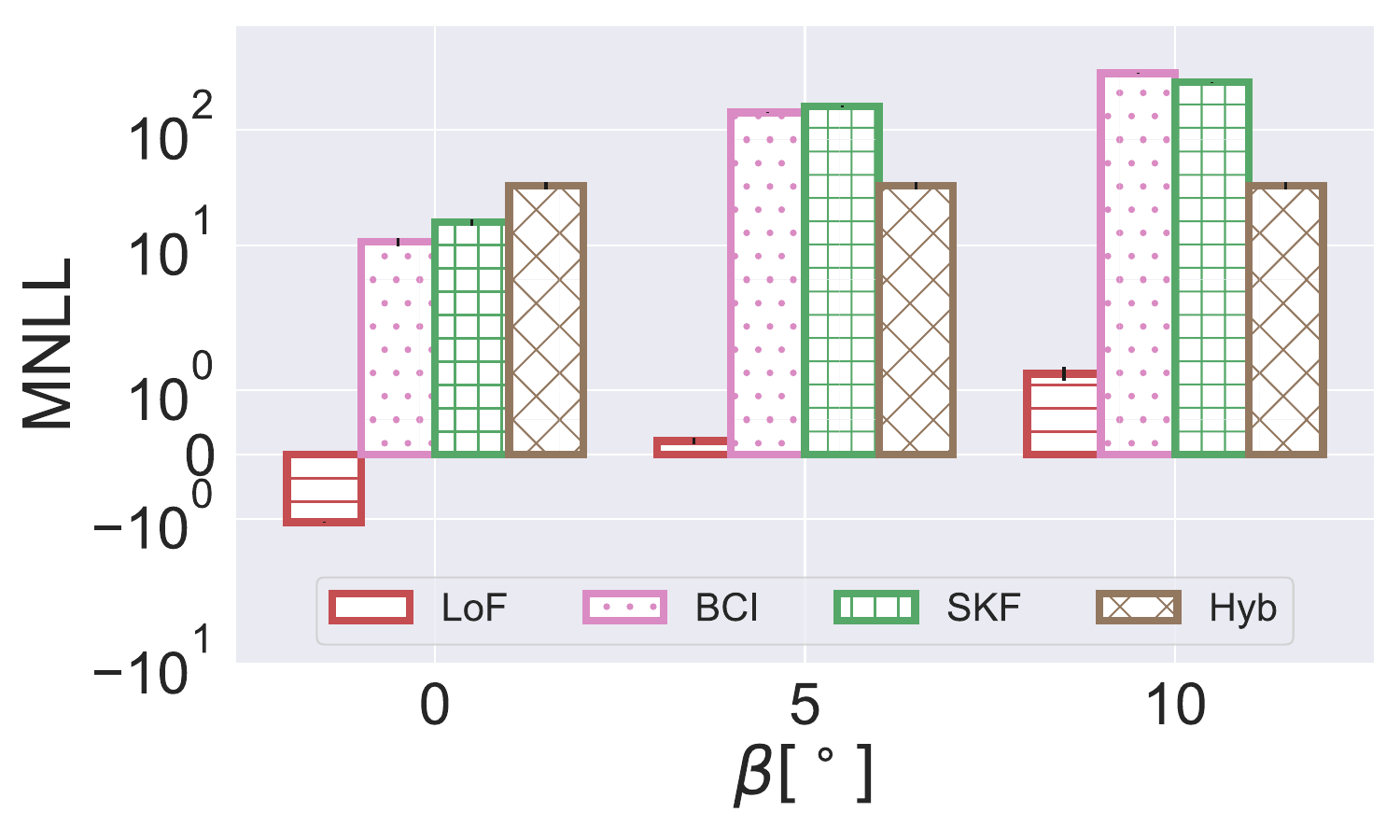}
  \end{minipage}\vspace{-1ex}
  \caption{Effect of Partial Known Observation Model.}\vspace{-1ex}
  \label{fig:exp-pkom}
\end{figure}

\begin{figure}
  \centering
  \begin{minipage}{0.48\columnwidth}
    \centering
    \includegraphics[width=.95\linewidth]{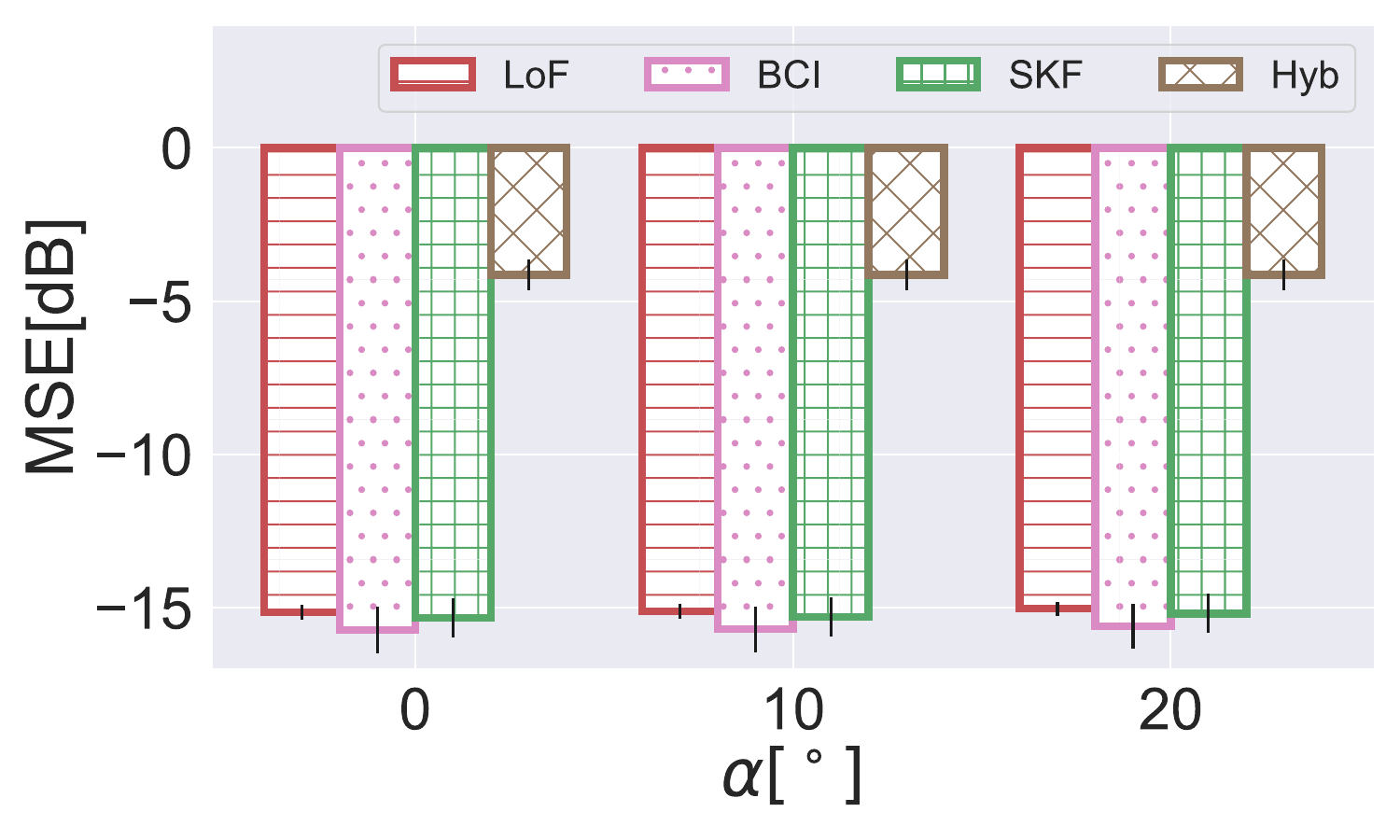}
  \end{minipage}
  \begin{minipage}{0.48\columnwidth}
    \centering
    \includegraphics[width=.95\linewidth]{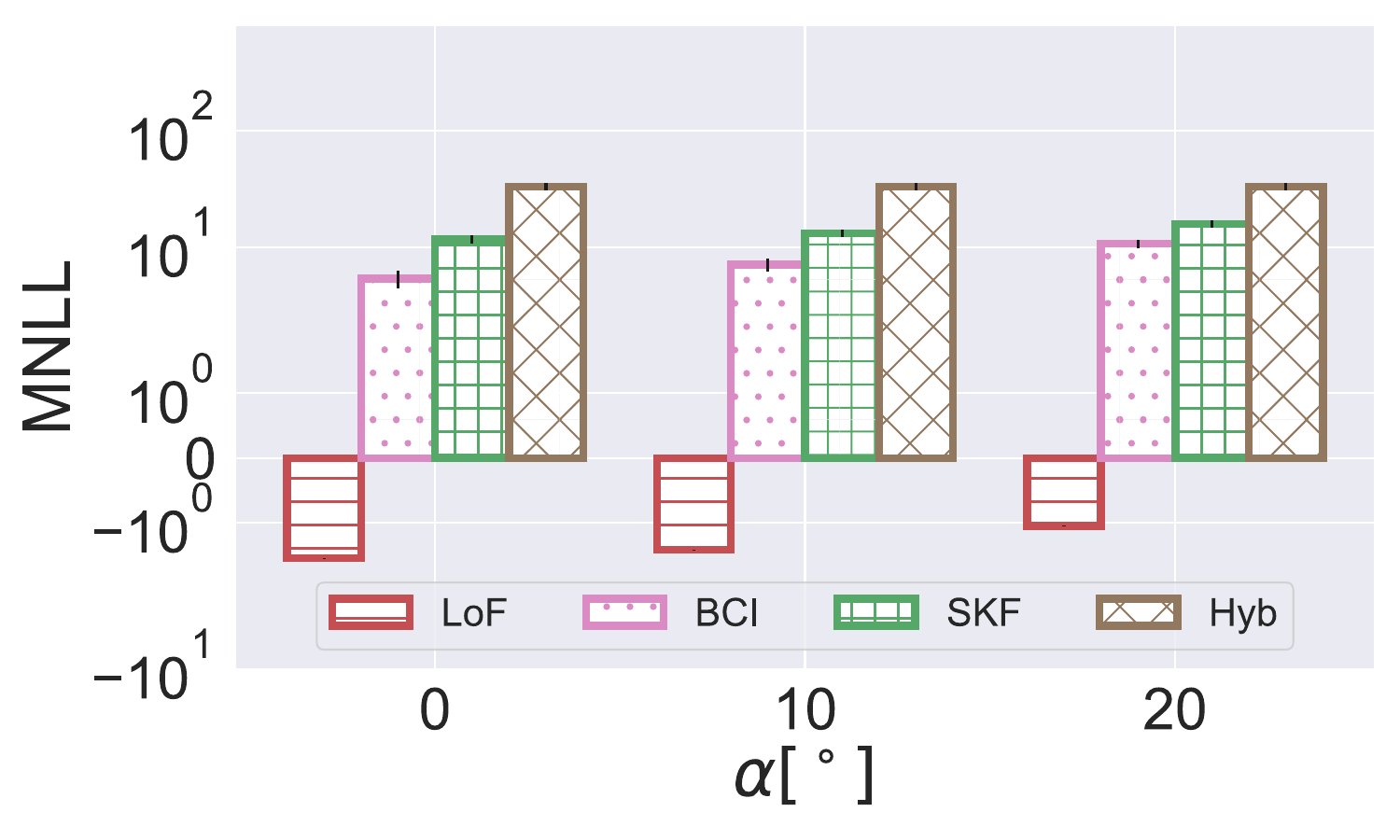}
  \end{minipage}\vspace{-1ex}
  \caption{Effect of Partial Known State-Evolution Model.}
  \label{fig:exp-pkem}
\end{figure}

\begin{figure}
  \centering
  \begin{minipage}{0.48\columnwidth}
    \centering
    \includegraphics[width=.95\linewidth]{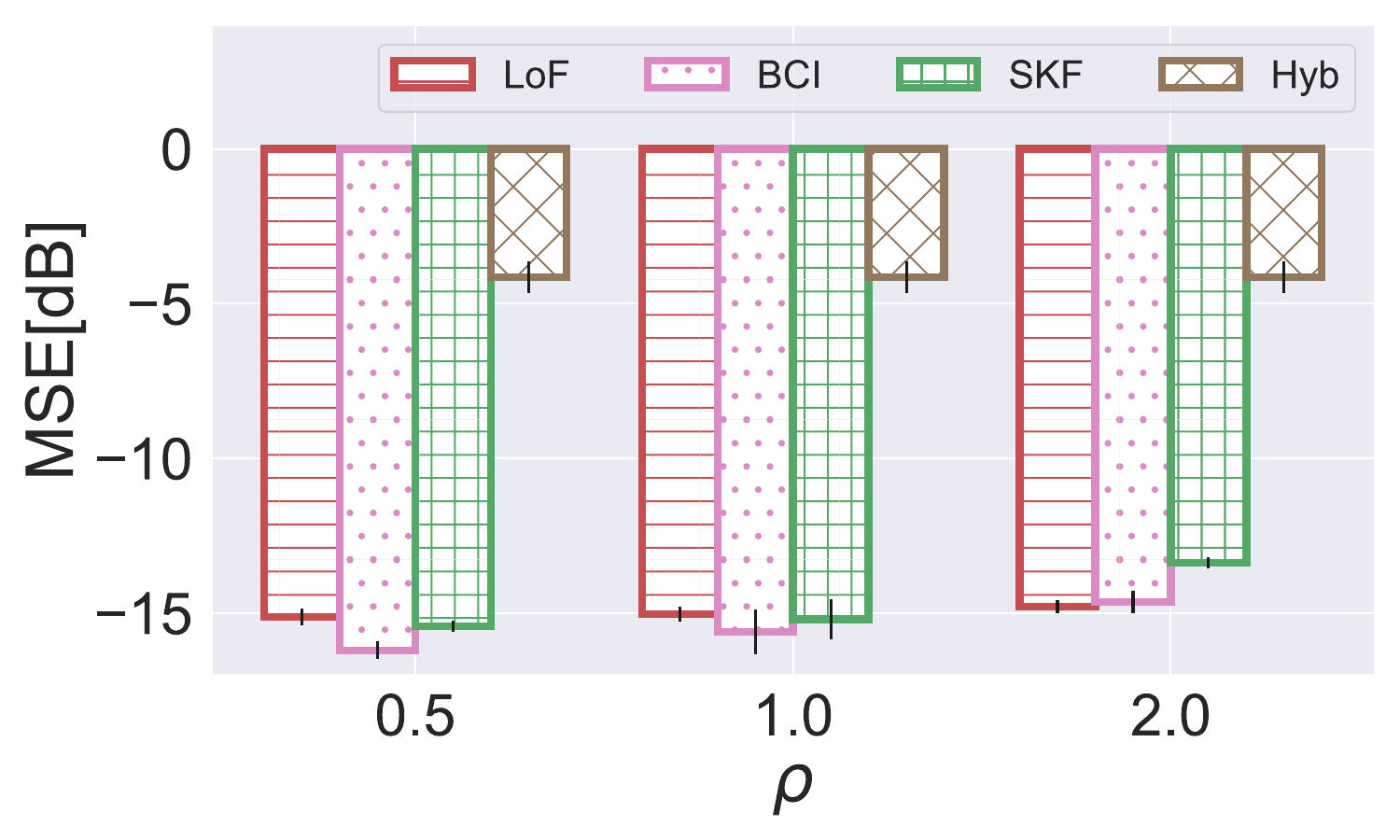}
  \end{minipage}
  \begin{minipage}{0.48\columnwidth}
    \centering
    \includegraphics[width=.95\linewidth]{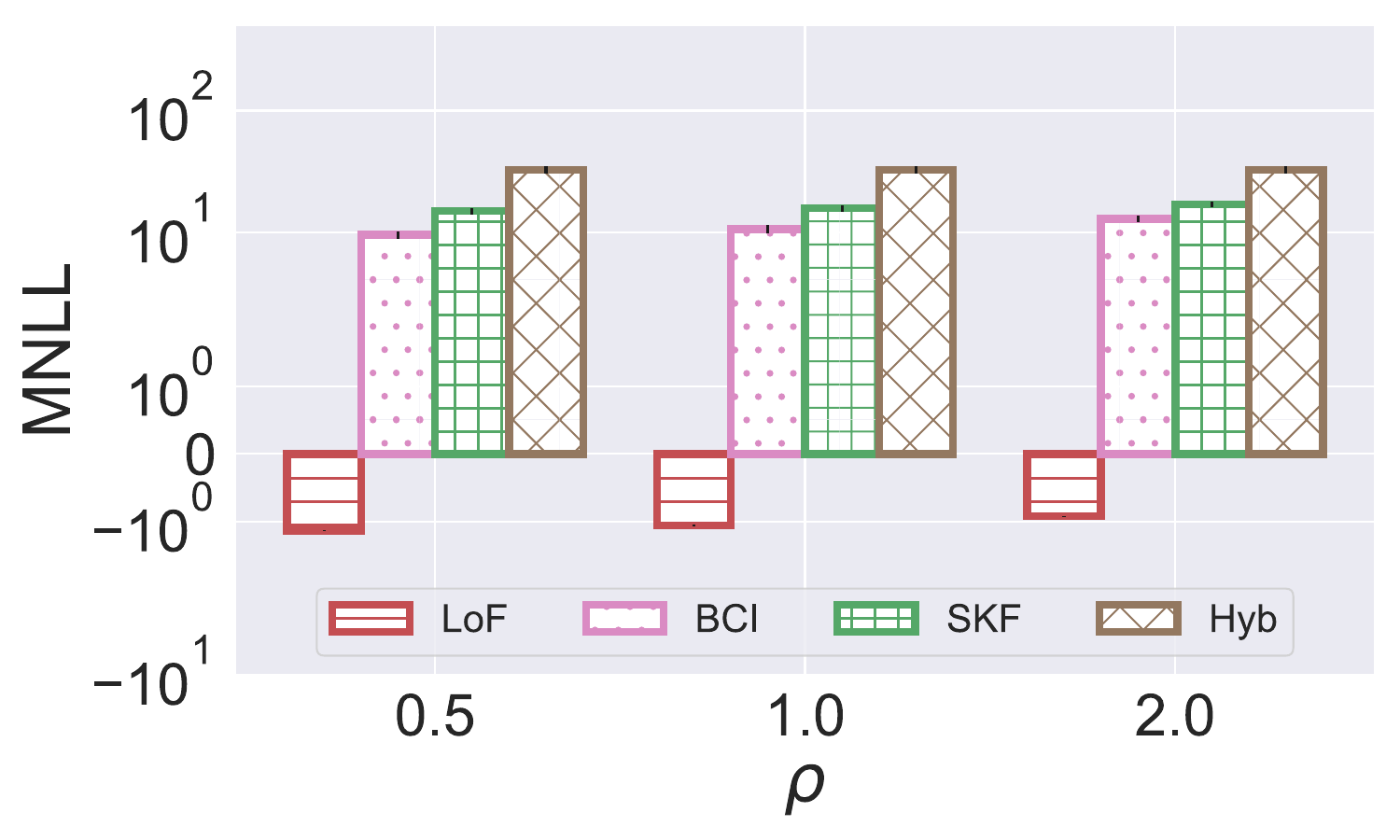}
  \end{minipage}\vspace{-1ex}
  \caption{Effect of Measurement Noise.}\vspace{-1ex}
  \label{fig:exp-mn}
\end{figure}

\subsubsection{Ablation Study}


\begin{table}[hb]
\begin{center}\footnotesize
\caption{Ablation Study}\label{exp:performance-ablation}
\begin{tabular}{cccc}
\toprule
 & \multicolumn{2}{c}{State} & \multicolumn{1}{c}{Uncertainty}  \\ \cmidrule(lr){2-3}\cmidrule(lr){4-4}
 & MSE[dB] & FG(\%) & MNLL \\ \hline
 LoF-M &   -10.02 ± 0.33       &  83.0 ± 0.1         &   38.95 ± 2.14       \\
LoF-TM       & -11.30 ± 0.07          & 87.3 ± 0.6          & 1.37 ± 0.08     \\
LoF+sm           & -11.33 ± 0.07          & 87.4 ± 0.6          & 1.53 ± 0.09    \\
LoF+$\tau_{0.5}$           & -11.33 ± 0.07          & 87.4 ± 0.6          & 1.50 ± 0.08    \\
LoF          &\textbf{-11.40 ± 0.06} & \textbf{87.6 ± 0.7} & \textbf{1.26 ± 0.11} \\
\bottomrule
\end{tabular}
\end{center}
\vspace{-2ex}
\end{table}
\normalsize

{In Table~\ref{exp:performance-ablation}, we perform the ablation study with four variants: \textsf{LoF}-T without TSM, \textsf{LoF}-TM without both TSM and $\mathsf{MLP}_{\theta}$, \textsf{LoF}+sm by using the soft medoid (SM) instead of the proposed TSM, and \textsf{LoF}+$\tau_{0.5}$ by setting a fixed decay factor $\tau = 0.5$. (1) When comparing \textsf{LoF}-T and \textsf{LoF}-TM, we find that using the learner $\mathsf{MLP}_{\theta}$, which generates local weights, significantly improves fusion performance.
(2) The results among \textsf{LoF}+sm, \textsf{LoF}-TM and \textsf{LoF} reveal that the introduction a Soft Medoid and TSM 
mitigates the noise effects during the fusion process.
(3) The comparison of \textsf{LoF}+$\tau_{0.5}$, \textsf{LoF}+sm and  \textsf{LoF} indicates that the proposed distance in TSM no matter with the fixed or adaptive decay ratio can improve the fusion. Particularly, the adaptive decay factor enhances the performance of TSM compared to the original Soft Medoid.}

\subsubsection{Parameter Sensitivity}

\begin{figure}
  \centering
  \begin{minipage}{0.48\columnwidth}
    \centering
    \includegraphics[width=.95\linewidth]{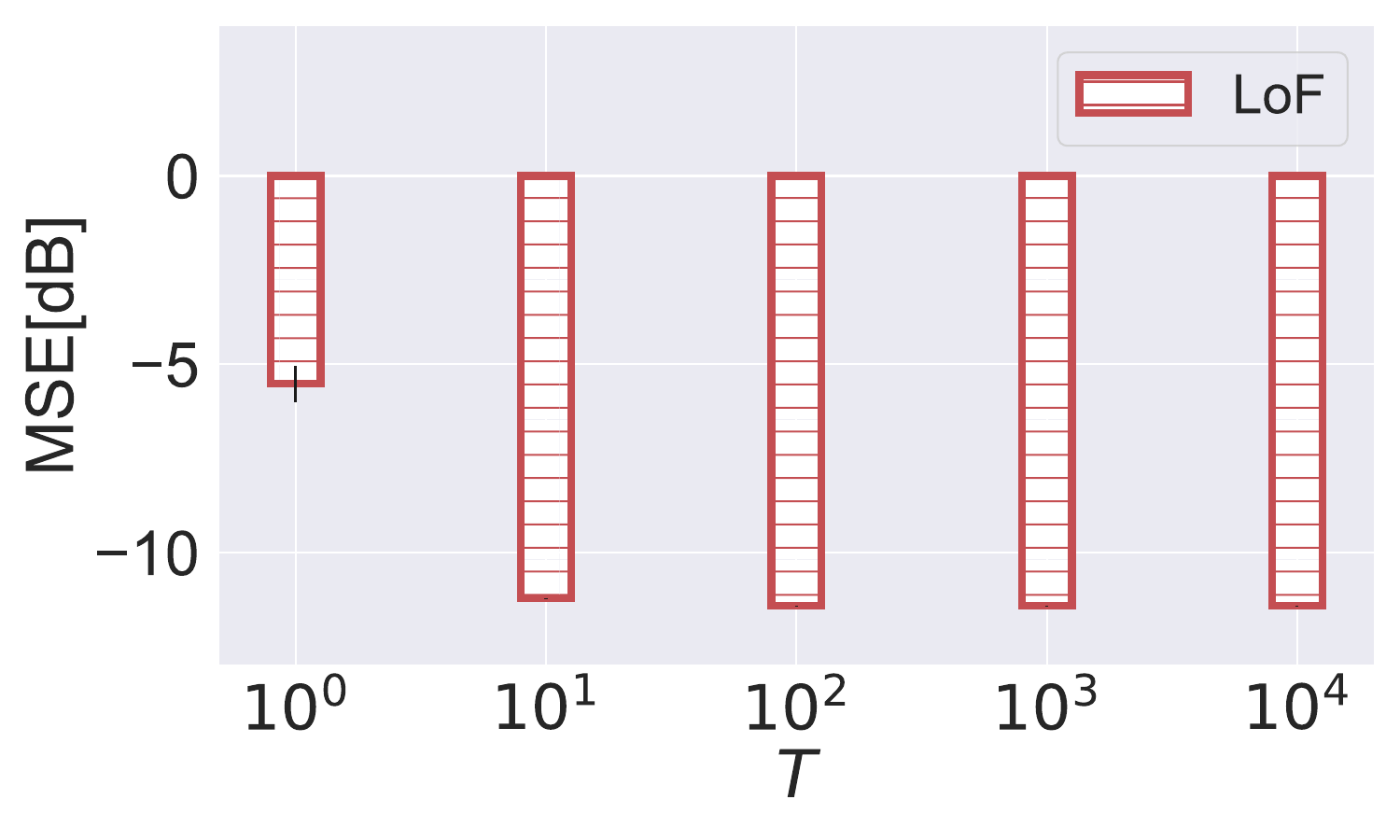}
  \end{minipage}
  \begin{minipage}{0.48\columnwidth}
    \includegraphics[width=0.95\columnwidth]{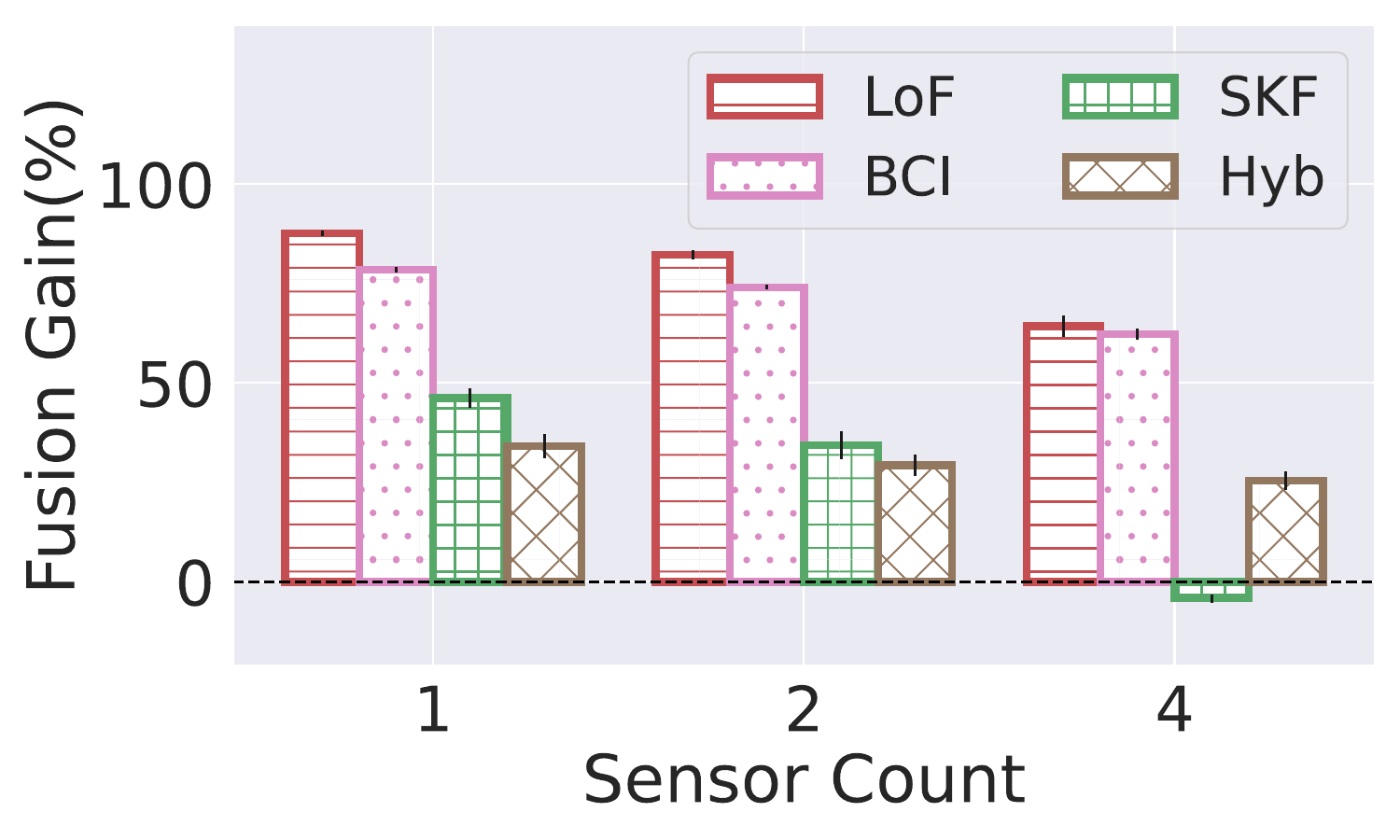}
        \end{minipage}\vspace{-1ex}
  \caption{Effect of (a) Temperature $\mathsf{T}$, (b) Disturbed sensors.}\vspace{-1ex}
  \label{fig:exp-sen}
\end{figure}



In this experiment, we study the effect of two parameters: the temperature $\mathsf{T}$ in TSM and the initialization value of the decay ratio $\tau^{k, i}_{t}$. In Fig.~\ref{fig:exp-sen} (a), a smaller temperature $\mathsf{T}$ on the overall leads to lower MSE errors and next remains stable. Intuitively, a too low temperature $\mathsf{T}$ means that TSM ignores all except the medoid, and otherwise a high temperature prefers to the non-robust mean. From this figure, we thus set $\mathsf{T} = 100$ in our evaluation by default. In addition, we examine the effect of the initialization value of the decay factor $\tau^{k, i}_{t}$. Due to space limit, we do not plot the associated figure. Our findings show that the initial value of $\tau^{k, i}_{t}$ has trivial effect on performance. It is mainly because the proposed adaptive decay factor mechanism can effectively mask the effect of such an initialization value.

\subsubsection{Disturbance Study}
Fig.~\ref{fig:exp-sen}(b) shows the effect of \textit{the number of disturbed sensors}. When this number grows, \textsf{LoF} and BCI perform well and yet the performance of SKF becomes significantly worse. It is mainly because both \textsf{LoF} and BCI employ the two-stage fusion and instead SKF adopts the fully centralized fusion. This experiment demonstrates that the two-stage scheme is rather robust. Compared to BCI, \textsf{LoF} performs still better mainly due to the adopted TSM against various disturbed sensors.

{Next, Table~\ref{exp:perturb-dist} evaluates the effect of two disturbance patterns. Recall that in the default pattern of \S \ref{sec5.13}, a selected sensor suffers from the disturbance from a specific time step until the final step. Here, the \emph{random} pattern shares the same disturbance strength as the default one, yet each sensor has a 25\% probability of being disturbed at every time step. In the \emph{strong}  pattern, we double the disturbance strength of the default one but with the same disturbance occurrence time steps. {By comparing the results in Table~\ref{exp:perturb-dist} and those in Table~\ref{exp:performance-CD} (with the `4a2t' setting), we find that \textsf{LoF} in the random pattern 
performs even better and instead the performance of three competitors becomes worse. This is mainly because the developed adaptive decay mechanism exploits the 2nd-order derivative to well adapt target state outliers. For the strong pattern, all four approaches degrade. Yet, \textsf{LoF} still performs the best. This experiment demonstrates that \textsf{LoF} performs well to be robust with various disturbance patterns.}

\begin{table}[tbp]
\begin{center}\footnotesize
\caption{Effect of Disturbance Patterns.}\label{exp:perturb-dist}
\begin{tabular}{ccccc}
\toprule
& & \multicolumn{2}{c}{State} & \multicolumn{1}{c}{Uncertainty}  \\ \cmidrule(lr){3-4}\cmidrule(lr){5-5}
& & MSE[dB] & FG(\%) & MNLL \\ \hline
 & BCI &    -8.54 ± 0.21       &  76.3 ± 1.1         &   299.49± 2.39       \\
Random &SKF       & -4.85 ±0.54          & 43.3 ± 12.5          & 261.26 ± 1.00     \\
Pattern &Hyb           & -3.81 ± 0.50          & 29.6 ± 1.1          & 37.04 ± 2.46    \\
&LoF          &\textbf{-11.89 ± 0.01} & \textbf{89.0 ± 1.1} & \textbf{0.56 ± 0.01} \\
\midrule
 & BCI &    -8.01 ± 0.26       &  87.8 ± 0.3         &   303.59 ± 2.52       \\
Strong &SKF       & -3.65 ±0.34          & 66.7 ± 1.5          & 241.59 ± 2.25     \\
Pattern &Hyb           & -3.80 ± 0.44         & -67.8 ± 2.1          & 37.70 ± 2.09    \\
&LoF          &\textbf{-8.88 ± 0.24} & \textbf{90.0 ± 0.2} & \textbf{1.64 ± 0.11} \\

\bottomrule
\end{tabular}
\end{center}
 \vspace{-2ex}
\end{table}
\normalsize


\subsubsection{Collaborative Detection} Finally, we evaluate the benefit of our work \textsf{LoF} on collaborative detection tasks. To this end, we replace the original SKF component of the state-of-the-art collaborative detection framework \cite{hsu2021scalable} by our \textsf{LoF}. In this way, the multi-agent reinforcement learning (MARL) algorithm in this  collaborative detection framework takes the fusion state estimate, including the mean and variance (uncertainty), as the input state to control the movement of mobile agents. We first train the entire algorithm with SKF and \textsf{LoF} in a `4a4t' environment, and next test it in the three `4a4t', `10a10t', and `50a50t' settings for generalization study.

In Fig.~\ref{fig:exp-as}(a), \textsf{LoF} consistently leads to smaller mean errors and less uncertainty (smaller variance width) than SKF, even when the number of agents and targets grows.  Intuitively, a higher number of agents lead to 
more local estimates to perform fusion and have chance to result in a lower MSE [dB]. Meanwhile,more targets inversely lead to a higher errors. Consequently, when the negative impact of additional targets outweighs the positive benefits caused by agents, the overall MSE [dB] increases. In addition, {Fig.~\ref{fig:exp-as}(b) gives the detection ratios of \textsf{LoF} and SKF. Again, \textsf{LoF} consistently outperforms SKF with a higher detection ratio and lower variance across different numbers of agents and targets, demonstrating high  generalization capability.} 



\begin{figure}[tbh]
  \centering
  \begin{minipage}{0.48\columnwidth}
    \centering
    \includegraphics[width=.95\linewidth]{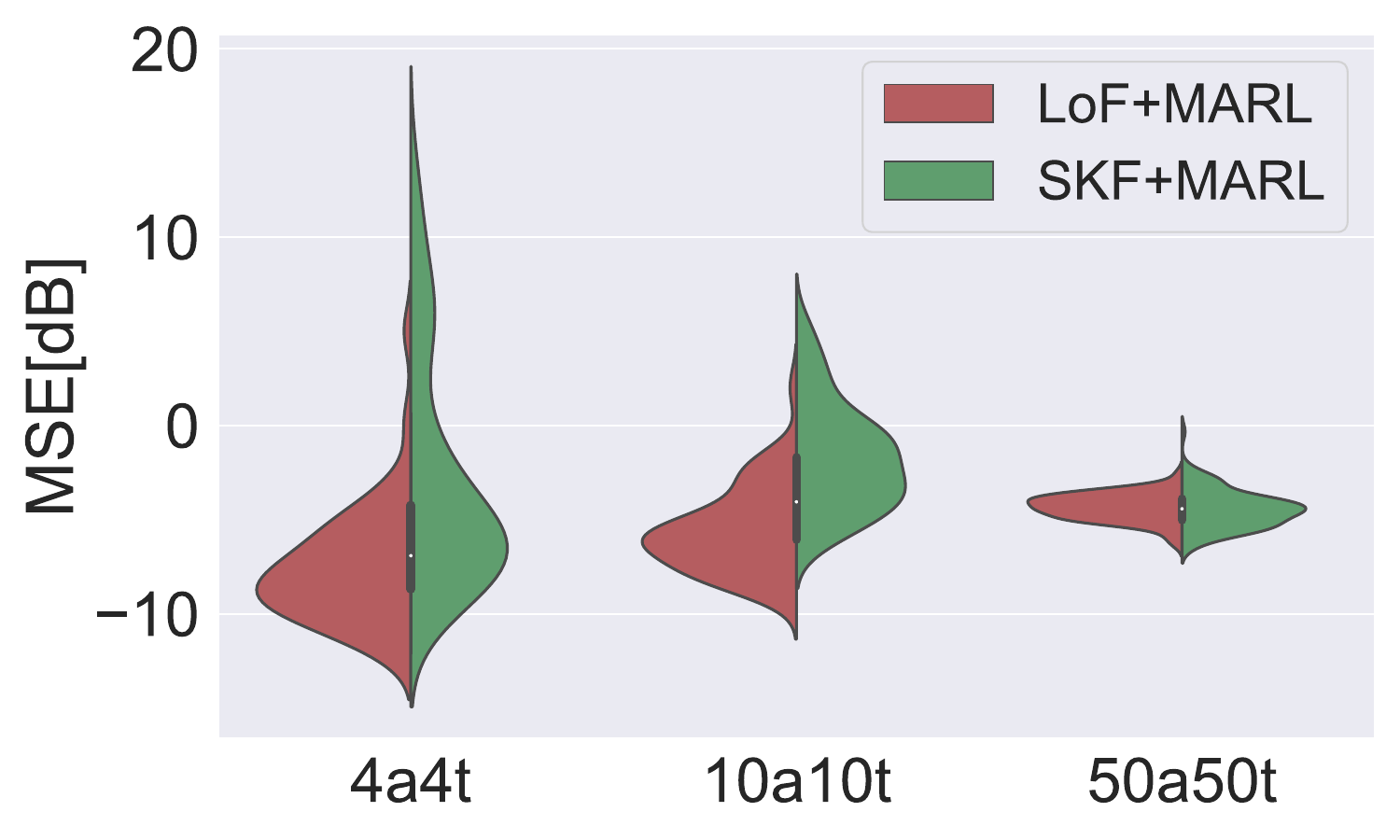}
  \end{minipage}
  \begin{minipage}{0.48\columnwidth}
    \includegraphics[width=0.95\columnwidth]{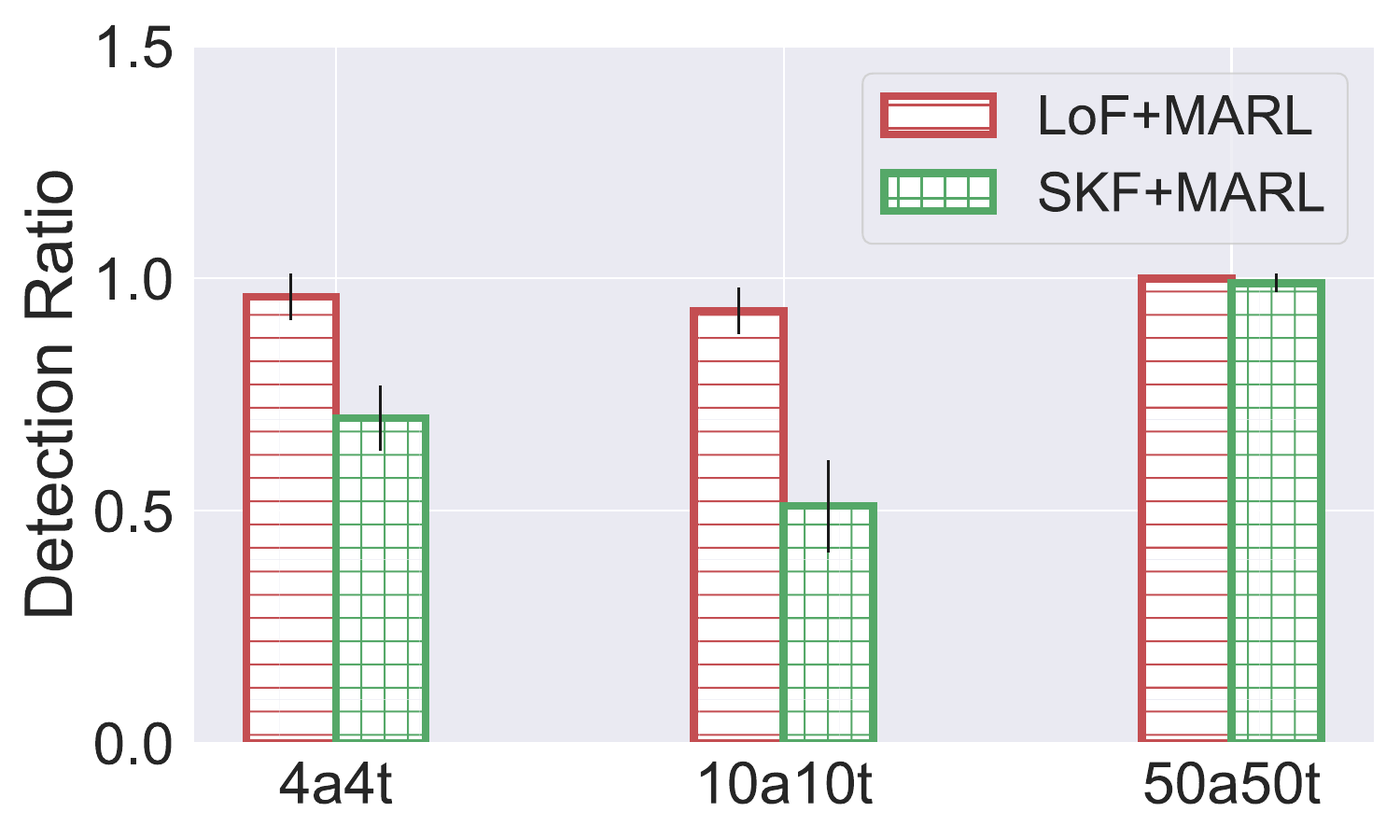}
        \end{minipage}\vspace{-1ex}
  \caption{Collaborative detection. (a) MSE and (b) detection ratio.}\vspace{-1ex}
\label{fig:exp-as}
\end{figure}

%% file: 06-conclusion.tex
\section{Conclusion}\label{sec7}
In this paper, we propose a two-stage collaborative fusion framework \textsf{LoF} to offer robust fusion in a partially known multi-agent environment. In the framework \textsf{LoF}, the local state estimator provides the probability distribution of target state mean and uncertainty, and the centralized robust fusion exploits the developed time-series Soft Medoid scheme (TSM) to perform robust fusion on the state probability distribution. Our extensive evaluation result in a collaborative detection simulation environment demonstrates the superiority of \textsf{LoF}. As future work, we are interested in the fusion scheme over heterogeneous observations, e.g., by on-board LIDAR and thermal imaging sensors, and perform real-world test.

%% file: 07-append.tex
\section*{Appendix}

\textbf{2. Proofs of Theorems 1 and 2}

\textit{2.1 Proof of Eqs.~(\ref{eq:center-fuse}) and~(\ref{eq:local-estimate}) in Theorem~\ref{th:fusion-problem}}.
\begin{proof}
To simplify the equations, the index $j$ of the target will be omitted in all subsequent derivations. Using the law of total probability, we derive the collaborative state estimation problem as follows:

\begin{equation}\small
p(\mathbf{x}_{t}|\mathcal{Y}_{1:t})  = \sum_{i=1}^{I}p(\mathbf{x}_{t},i|\mathcal{Y}_{1:t})
\label{eq:t1-p1}
\end{equation}

According to the Bayes' theorem, we have

\begin{equation}\small
\begin{array}{rl}
p(\mathbf{x}_{t},i|\mathcal{Y}_{1:t}) & = p(\mathbf{x}_{t}|i,\mathcal{Y}_{1:t})p(i|\mathcal{Y}_{1:t}) \\
& = p(\mathbf{x}_{t}|\mathbf{y}_{t}^{i},\mathcal{Y}_{1:t-1}) p(i|\mathcal{Y}_{1:t}) \\
\end{array}
\label{eq:t1-p2}
\end{equation}

Combining Eq.~(\ref{eq:t1-p1}) and Eq.~(\ref{eq:t1-p2}), we obtain

\begin{equation}\small
p(\mathbf{x}_{t}|\mathcal{Y}_{1:t}) = \sum_{i=1}^{I}p(\mathbf{x}_{t}|\mathbf{y}_{t}^{i},\mathcal{Y}_{1:t-1}) p(i|\mathcal{Y}_{1:t}) \\
\label{eq:t1-p3}
\end{equation}

$p(\mathbf{x}_{t}|\mathbf{y}_{t},\mathcal{Y}_{1:t-1})$ denotes the $i$-th agent's estimate of the target state based on its current observation and the historical estimates of all agents, $p(i|\mathcal{Y}_{1:t})$ is the fusion weight of $i$-th agent. However, directly collecting historical estimates from all agents is costly. Therefore, we further derive Eq.~(\ref{eq:t1-p3}). According to Bayes' theorem, we have
\begin{equation}\small
\begin{array}{rl}
    p(\mathbf{x}_{t}|\mathbf{y}_{t}^{i},\mathcal{Y}_{1:t-1}) & = \frac{p(\mathbf{x}_{t},\mathbf{y}_{t}^{i}|\mathcal{Y}_{1:t-1})}{p(\mathbf{y}_{t}^{i}|\mathcal{Y}_{1:t-1})} \\
    & = \frac{p(\mathbf{y}_{t}^{i}|\mathbf{x}_{t}, \mathcal{Y}_{1:t-1})p(\mathbf{x}_{t}|\mathcal{Y}_{1:t-1})}{p(\mathbf{y}_{t}^{i}|\mathcal{Y}_{1:t-1})} \\
\end{array}
\end{equation}
without loss of generality, we follow the traditional Bayesian Filter and assume the new measurement $\mathbf{y}_{t}^{i}$ is independent of the previous measurements $\mathcal{Y}_{1:t-1}$, and we have
\begin{equation}\small
    p(\mathbf{x}_{t}|\mathbf{y}_{t}^{i},\mathcal{Y}_{1:t-1}) = \frac{p(\mathbf{y}_{t}^{i}|\mathbf{x}_{t})p(\mathbf{x}_{t}|\mathcal{Y}_{1:t-1})}{p(\mathbf{y}_{t}^{i}|\mathcal{Y}_{1:t-1})}
\label{eq:t1-p4}
\end{equation}
here, Eq.~(\ref{eq:t1-p4}) is similar to the KF's update step. According to the derivation of KF's update step in~\cite{masnadi2019step}, we have
\begin{subequations} \label{eq:t1-p5-6}
    \begin{align}
     p(\mathbf{x}_{t}|\mathcal{Y}_{1:t-1})    & =  \int p(\mathbf{x}_{t}|\mathbf{x}_{t-1})p(\mathbf{x}_{t-1}|\mathcal{Y}_{1:t-1}) \mathrm{d}\mathbf{x}_{t-1}  \label{eq:t1-p5}  \\
     p(\mathbf{y}_{t}^{i}|\mathcal{Y}_{1:t-1}) & = \int p(\mathbf{y}_{t}^{i}|\mathbf{x}_{t})p(\mathbf{x}_{t}|\mathcal{Y}_{1:t-1}) \mathrm{d}\mathbf{x}_{t}  \label{eq:t1-p6} 
    \end{align}
\end{subequations}
where $p(\mathbf{x}_{t}|\mathbf{x}_{t-1})$ indicates the state-evolution model and $p(\mathbf{y}_{t}^{i}|\mathbf{x}_{t})$ represents the observation model. We can find that Eq.~(\ref{eq:t1-p5}) is similar to the KF's prediction step, but KF's prediction step is based on the historical observations of a single sensor $\mathbf{y}_{1:t-1}$, while Eq.~(\ref{eq:t1-p5}) is based on the historical observations of all sensors $\mathcal{Y}_{1:t-1}$. To this end, we can conclude that $p(\mathbf{x}_{t}|\mathbf{y}_{t}^{i},\mathcal{Y}_{1:t-1})$ can be represent as a function with inputs $\mathbf{y}_{t}^{i}$ and $p(\mathbf{x}_{t-1}|\mathcal{Y}_{1:t-1})$:
\begin{equation}
    \psi(p(\mathbf{x}_{t-1}|\mathcal{Y}_{1:t-1}), \mathbf{y}_{t}^{i})
\end{equation}
and the function $\psi(\cdot)$ can be realized as a KF estimator. 






\end{proof}

\textit{2.2 Proof of Eq.~(\ref{eq:lweight}) in Theorem~\ref{th:fusion-problem}}.
\begin{proof}
We derive $p(i|\mathcal{Y}_{1:t})$ according to the law of total probability and Bayes' theorem as follows:
\begin{equation}
    \begin{array}{rl}
    p(i|\mathcal{Y}_{1:t})  & = p(i|\mathcal{Y}_{t},\mathcal{Y}_{1:t-1})  \\
         & =\frac{p(i,\mathcal{Y}_{t}|\mathcal{Y}_{1:t-1})}{p(\mathcal{Y}_{t}|\mathcal{Y}_{1:t-1})} \\
         & = \frac{p(\mathcal{Y}_{t}|i,\mathcal{Y}_{1:t-1})p(i|\mathcal{Y}_{1:t-1})}{\sum_{i^{\prime}=1}^{I}p(\mathbf{Y}_{t},i^{\prime}|\mathbf{Y}_{1:t-1})} \\
         & = \frac{p(\mathbf{y}_{t}^{i}|\mathcal{Y}_{1:t-1})p(i|\mathcal{Y}_{1:t-1})}{\sum_{i^{\prime}=1}^{I}p(\mathcal{Y}_{t}|i^{\prime},\mathcal{Y}_{1:t-1})p(i^{\prime}|\mathcal{Y}_{1:t-1})} \\
         & = \frac{p(\mathbf{y}_{t}^{i}|\mathcal{Y}_{1:t-1})p(i|\mathcal{Y}_{1:t-1})}{\sum_{i^{\prime}=1}^{I}p(\mathbf{y}_{t}^{i^{\prime}}|\mathcal{Y}_{1:t-1})p(i^{\prime}|\mathcal{Y}_{1:t-1})}
    \end{array}
\label{eq:conf-level}
\end{equation}

That is,
\begin{equation}
p(i|\mathcal{Y}_{1:t}) = \phi ( p(\mathbf{y}_{t}^{i}|\mathcal{Y}_{1:t-1})p(i|\mathcal{Y}_{1:t-1}) )
\label{eq:t1-p7}
\end{equation}
where $\phi(\cdot)$ indicates the normalization process.
\end{proof}


\textbf{3. Proof of Theorem~\ref{th:o-weight}}
\begin{proof}
According to Eq.~(\ref{eq:t1-p7}), we can recursively calculate weight $p(i|\mathcal{Y}_{1:t})$ with the help of $p(\mathbf{y}_{t}^{i}|\mathcal{Y}_{1:t-1})$. According to the law of total probability and Bayes' theorem, we have
\begin{equation}
    \begin{array}{rl}
     p(\mathbf{y}_{t}^{i}|\mathcal{Y}_{1:t-1})    & =  \int p(\mathbf{y}_{t}^{i}, \mathbf{x}_{t}|\mathcal{Y}_{1:t-1}) \mathrm{d}\mathbf{x}_{t}   \\
      & = \int p(\mathbf{y}_{t}^{i}|\mathbf{x}_{t})p(\mathbf{x}_{t}|\mathcal{Y}_{1:t-1}) \mathrm{d}\mathbf{x}_{t} \\
    \end{array}
\label{eq:ino-dist}
\end{equation}

From KF, we have
\begin{equation}
    \begin{array}{rl}
    p(\mathbf{y}_{t}^{i}|\mathbf{x}_{t}) & = \textsf{PDF}(\mathbf{y}_{t}^{i}|\langle\mathbf{H}_{i}\mathbf{x}_{t},\mathbf{R}_{i}\rangle)\\
    p(\mathbf{x}_{t}|\mathcal{Y}_{1:t-1}) & = \textsf{PDF}(\mathbf{x}_{t}|\langle\mathbf{x}_{t|t-1},\mathbf{\Sigma}_{t|t-1}\rangle)\\
    \end{array}
\label{eq:ino-prob}
\end{equation}
where $\mathbf{H}_{i}$ and $\mathbf{R}_{i}$ is the parameter of SS model and $\textsf{PDF}(\cdot)$ indicates the probability density function of the Gaussian distribution. Combining Eq.~(\ref{eq:ino-dist}) and Eq.~(\ref{eq:ino-prob}), we have
\begin{equation} \scriptsize
    \begin{array}{rl} 
     p(\mathbf{y}_{t}^{i}|\mathcal{Y}_{1:t-1})  &  = \int \textsf{PDF}(\mathbf{y}_{t}^{i}|\langle\mathbf{H}_{i}\mathbf{x}_{t},\mathbf{R}_{i}\rangle)\textsf{PDF}(\mathbf{x}_{t}|\langle\mathbf{x}_{t|t-1},\mathbf{\Sigma}_{t|t-1}\rangle) \mathrm{d}\mathbf{x}_{t} \\
      & = \textsf{PDF}(\mathbf{y}_{t}^{i}|\langle\mathbf{H}_{i}\mathbf{x}_{t|t-1},\mathbf{H}_{i}\mathbf{\Sigma}_{t|t-1}\mathbf{H}_{i}^{T}+\mathbf{R}_{i}\rangle) \\
      & = \textsf{PDF}(\mathbf{y}_{t}^{i}|\langle\hat{\mathbf{y}}_{t}^{i},\mathbf{S}_{t|t-1}^{i}\rangle) \\
      & = \textsf{PDF}(\Delta \mathbf{y}_{t}^{i}|\langle \mathbf{0},\mathbf{S}_{t|t-1}^{i}\rangle) \\
    \end{array}
\normalsize
\end{equation}
where $\mathbf{S}_{t|t-1}^{i}$ is the covariance matrix of \emph{innovation} (i.e., the residual $\Delta \mathbf{y}_{t}^{i}=\mathbf{y}_{t}^{i}-\hat{\mathbf{y}}_{t}^{i}$).

\end{proof}